\documentclass[10pt,twocolumn,letterpaper]{article}

\usepackage[pagenumbers]{cvpr} 
\usepackage{tabularray}
\usepackage[ruled,vlined,linesnumbered,ruled]{algorithm2e}
\usepackage{multicol}
\usepackage{multirow}
\usepackage{graphicx}
\usepackage{amsmath}
\usepackage{amssymb}
\usepackage{booktabs}
\usepackage{colortbl}
\usepackage{array}
\usepackage{float}  
\usepackage{hhline}
\usepackage{pifont}
\usepackage{arydshln}

\newcommand{\codename}{{LAMP}\xspace}

\definecolor{cvprblue}{rgb}{0.21,0.49,0.74}
\usepackage[pagebackref,breaklinks,colorlinks,allcolors=cvprblue]{hyperref}

\def\paperID{***} 
\def\confName{CVPR}
\def\confYear{2026}

\title{LAMP: Lift Image-Editing as General 3D Priors for Open-world Manipulation}

\author{
Jingjing Wang$^{1}$\quad
Zhengdong Hong$^{1}$ \quad
Chong Bao$^{1}$  \\
Yuke Zhu$^{1}$ \quad
Junhan Sun$^{1}$ \quad
Guofeng Zhang$^{1}$\footnotemark[2] \\
 \small $^{1}$State Key Lab of CAD\&CG, Zhejiang University 
}

\begin{document}

\twocolumn[{%
\renewcommand\twocolumn[1][]{#1}%
\maketitle
\vspace{-2em}
\begin{center}
    \centering
    \includegraphics[width=1.0\linewidth, trim={0 0 0 0}, clip]{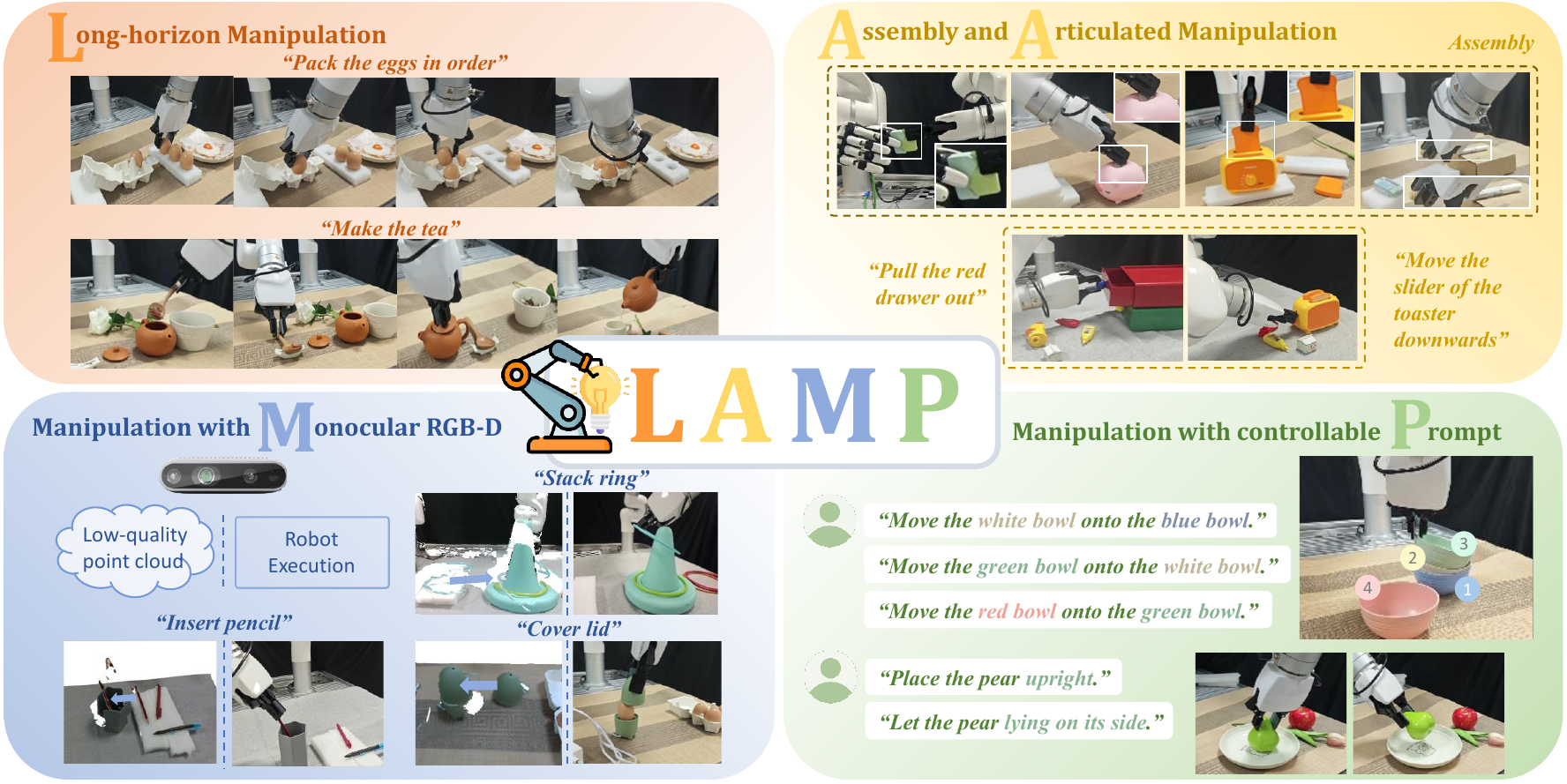}
    \vspace{-2.0em}
    \captionof{figure}{We propose \textbf{\codename}, which lifts image editing as general 3D priors, enabling open-world manipulation of diverse tasks from monocular RGB-D observations and promptable instructions.}
    \label{fig:teaser}
\end{center}%
}]

\maketitle
\begin{abstract}
Human-like generalization in open-world remains a fundamental challenge for robotic manipulation. 
Existing learning-based methods, including reinforcement learning, imitation learning, and vision-language-action-models (VLAs), often struggle with novel tasks and unseen environments. 
Another promising direction is to explore generalizable representations that capture fine-grained spatial and geometric relations for open-world manipulation. 
While large-language-model (LLMs) and vision-language-model (VLMs) provide strong semantic reasoning based on language or annotated 2D representations, their limited 3D awareness restricts their applicability to fine-grained manipulation.
To address this, we propose \codename, which lifts image-editing as 3D priors to extract inter-object 3D transformations as continuous, geometry-aware representations. 
Our key insight is that image-editing inherently encodes rich 2D spatial cues, and lifting these implicit cues into 3D transformations provides fine-grained and accurate guidance for open-world manipulation. 
Extensive experiments demonstrate that \codename delivers precise 3D transformations and achieves strong zero-shot generalization in open-world manipulation.
Project page: \href{https://zju3dv.github.io/LAMP/}{https://zju3dv.github.io/LAMP/}.
\vspace{-1.0em}
\end{abstract}
    
\section{Introduction}
\label{sec:intro}
Achieving human-like generalization in open-world robotic manipulation remains one of the ultimate goals for embodied intelligence. 
The challenge stems from the wide variety of task structures, levels of complexity, and temporal horizons.
Traditional methods typically rely on task-specific modeling of robot states~\cite{bicchi2000robotic,mordatch2012contact,rus1999hand}, which limits their generalizability.
Recent learning-based approaches like reinforcement learning (RL)~\cite{aytar2018playing,rajeswaran2018learning,xu2023unidexgrasp,hansenmodem,moerland2023model,handa2023dextreme, hong2025learning,kumar2023graph,zakka2022xirl}, imitation learning (IL)~\cite{jarrett2020strictly,reuss2023goal,seo2023trill,chi2025diffusion}, and VLAs~\cite{black2025pi_,brohan2022rt,zitkovich2023rt,kim2025openvla,li2024cogact,belkhale2024rt,o2024open,stoneopen} adopt a data-driven paradigm by training networks on various robot data.
But they struggle to handle novel tasks and environments that are entirely different, falling short in open-world manipulation.
To reach the goal, another strategy is to explore a generalizable representation for robotic manipulation in open worlds.

One promising direction for open-world manipulation is to leverage the spatial reasoning ability of LLMs~\cite{achiam2023gpt,team2024gemma,team2024gemini,bai2023qwen} and VLMs~\cite{liu2023visual,wang2024cogvlm,bai2025qwen2,guo2025seed1}. 
Some works~\cite{liang2023code,huang2023voxposer} leverage the code-generation capability of LLMs to represent manipulation as executable code segments from language instructions.
This representation effectively converts simple and concrete spatial expressions (e.g., ``move up", ``go left", ``1 meter away") into actionable code primitives, yet lacks perception of the actual scene and geometry due to the absence of visual grounding.
Other methods~\cite{huang2025rekep,huang2024copa,fangandliu2024moka,nasiriany2024pivot,pan2025omnimanip} instead represent manipulation as geometric relations (e.g. distance, parallelism, or perpendicularity) between annotated entities (e.g., keypoints or vectors) on 2D observations.
While effective for simple spatial reasoning, these explicit 2D annotations are fragile under noisy depth and viewpoint changes.
Despite their difference, both LLM- and VLM-based previous approaches ultimately rely on language-described explicit constraints that are inherently sparse and ambiguous in 3D space.
They struggle to express fine-grained geometric relations, such as relative rotations, contact geometry or precise alignment between interacting objects, which are essential for precise manipulation like assembly~\cite{chen2022neural}.
The core limitation stems from the discrete and symbolic nature of language, which makes it hard to capture continuous 3D spatial interactions.

To address this challenge, we seek a representation that captures continuous and geometry-aware spatial relations beyond discrete linguistic constraints and remains robust to viewpoint variations.
Inspired by assembly tasks, we adopt inter-object 3D transformations as a physically grounded representation for manipulation.
Such transformations naturally encode relative motion, contact geometry, and alignment between objects in 3D space.
However, obtaining accurate 3D priors remains nontrivial.
Video-~\cite{bharadhwaj2024track2act,bharadhwajgen2act,patel2025robotic} and 4D-generative models~\cite{zhen2025tesseract} provide a potential path to extract such priors, but currently they still suffer from severe visual inconsistency and incorrect functional understanding, while being computationally expensive.
We instead observe that image-editing models implicitly encode rich spatial priors in the 2D visual domain: how an object should move, rotate, or interact within a scene.
Moreover, due to their paired image supervision and object-consistent editing behavior, these models maintain strong subject consistency across edits.
This motivates our central question: \textit{Can we extract 3D priors for manipulation from image editing?}

We introduce \textbf{\codename} (\textbf{L}ift Im\textbf{A}ge-Editing as General 3D Priors for Open-World \textbf{M}ani\textbf{P}ulation). It lifts spatial clues in edited images into 3D inter-object transformations. Specifically, given a task instruction, we first perform image editing on the current observation to obtain an edited state. Using the current depth map from a RGB-D camera and single-view reconstruction~\cite{wang2025vggt}, we lift the current and edited states into their 3D coordinate frames, and compute their inter-object transformation by aligning the active and passive manipulation objects of the current frame to the edited frame.
This dense 3D transformation acts as a continuous geometric prior, encoding both spatial alignment and interaction intent.
We enhance robustness to depth noise with 2D-3D fused hierarchical point-cloud filtering, which retains only reliable partial geometry under viewpoint variation.
We further handle the potential inter-object scale inconsistency introduced by image-editing~\cite{cheng20253d} (e.g., object size change between the current and edited states) via scale alignment. 
Our main contributions are as follows:
\begin{itemize}
\item We propose \codename, which lifts image-editing into 3D general priors for manipulation and extracts precise inter-object 3D transformations from single-view RGB-D observations, enabling efficient open-world manipulation.
\item We provide an in-depth analysis of current VLM/LLM-based open world manipulation methods and demonstrate the superior generalization and robustness of our image-editing-lifted 3D priors in open world settings. 
\item Through extensive experiments, we demonstrate our method's strong zero-shot generalization across a diverse variety of real-world manipulation tasks.
\end{itemize}

\begin{figure*}[ht]
    \centering
    \includegraphics[width=1.0\linewidth]{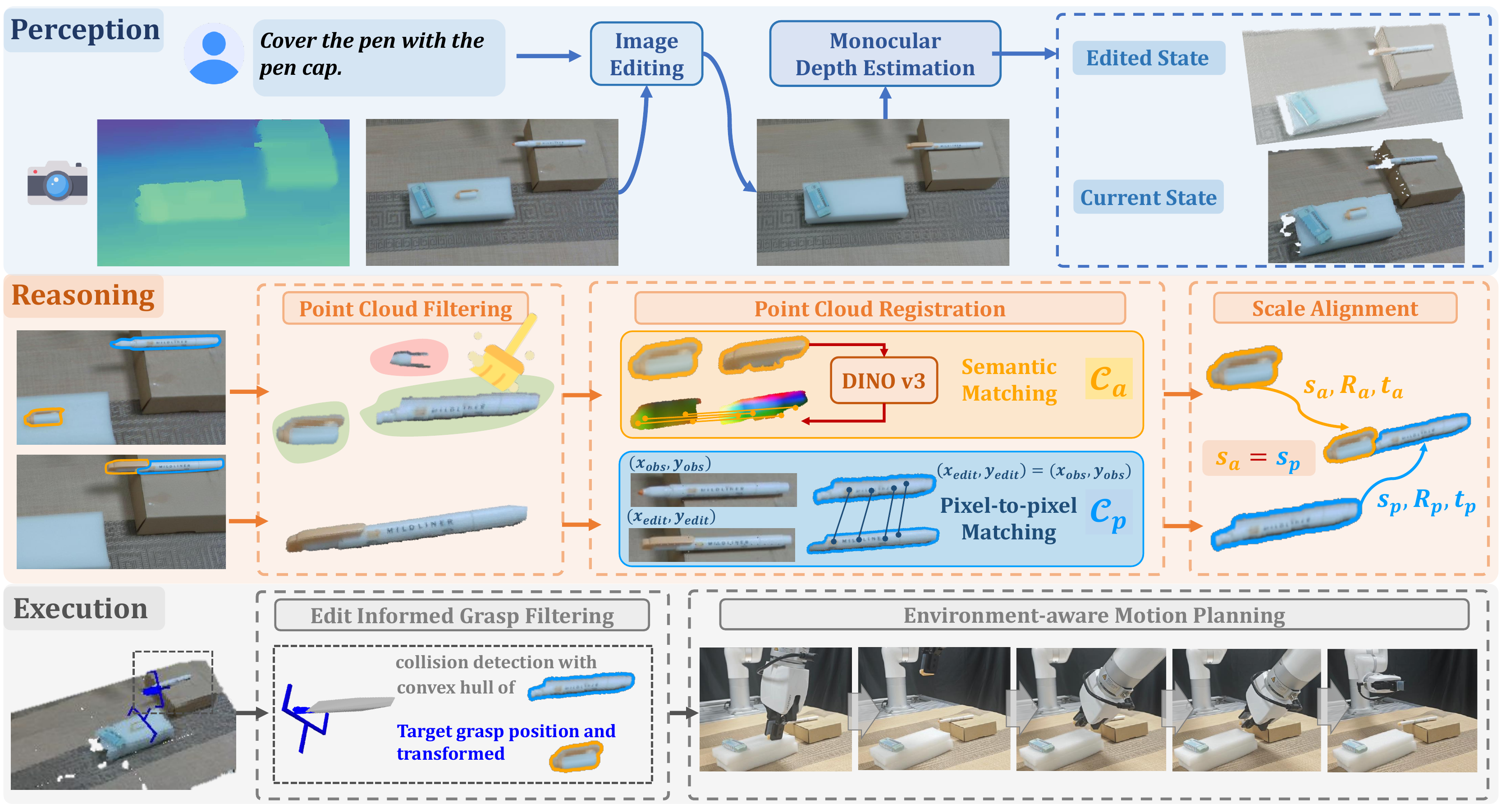}
    \vspace{-2.0em}
    \caption{\textbf{Overview.} Given the RGB-D observation and a language instruction, the Image-editing generates an edited state, which is used for registration to extract the inter-object transformation in reasoning stage. This transformation is converted into target pose for execution.}
    \label{fig:pipeline}
\vspace{-1.5em}
\end{figure*}

\section{Related Works}
\label{sec:related}

\noindent \textbf{General Representations for Robotic Manipulation.}
General representations are the key to achieving strong generalization in open-world manipulation. 
Traditional end-to-end policies typically employ neural networks to extract spatial features, learning dense neural descriptors as object-centric representations for downstream control~\cite{simeonov2022neural,heravi2023visuomotor,yuan2022sornet,sundaresan2020learning,simeonov2023se,chun2023local} to enable in-category generalization.
To tackle open-world manipulation, recent efforts construct structured visual inputs to prompt Vision-Language Models (VLMs) or Large Language Models (LLMs). These methods leverage visual foundation models to extract semantic keypoints~\cite{huang2025rekep, nasiriany2024pivot, fangandliu2024moka}, calculate projected motion vectors~\cite{huang2024copa}, or estimate explicit 3D poses~\cite{pan2025omnimanip}. 
While highly interpretable for reasoning, these explicit intermediate representations are often brittle under occlusion, viewpoint shifts, and depth noise, leading to unstable grounding across diverse scenes.
Another direction relies on template matching or regression networks to predict 6D object poses or bounding boxes as intermediate representations~\cite{tremblay2018deep}; however, such explicit pose estimation often struggles to generalize across unseen, out-of-distribution objects.
Another direction employs 3D flow as a motion representation. While earlier methods~\cite{eisner2022flowbot3d} relied on scarce synthetic 3D assets, recent works like FLIP~\cite{gaoflip} and Dream2Flow~\cite{dharmarajan2025dream2flow} leverage generative video priors to extract visual flow without manual annotations. 
Despite its flexibility, flow remains a local, point-wise description that lacks explicit structural grounding between interacting objects. This makes it difficult to reason about the precise $SE(3)$ constraints required for complex tasks like assembly.
Alternatively, several works focus on learning inter-object transformations as generalizable representations, particularly for assembly tasks~\cite{zhao2025anyplace,qi2025two,chen2022neural,wu2023leveraging,lu2023jigsaw,wangpuzzlefusion++,sun2025rectified}. 
However, these methods typically depend on complete 3D geometry inputs and require task-specific training.
To bypass these limitations, this work lifts 2D image-editing priors into robust 3D inter-object SE(3) transformations. This yields a spatially grounded representation that remains stable under real-world noise and partial, monocular observations.

\noindent \textbf{Foundation Models for Manipulation.}
Foundation models increasingly leverage large-scale vision-language priors to facilitate embodied reasoning and task planning~\cite{kawaharazuka2024real, firoozi2025foundation}.
VLM- and LLM-based methods bridge high-level reasoning with low-level execution by extracting spatial cues such as 3D action maps~\cite{huang2023voxposer}, relational keypoints~\cite{huang2025rekep}, or interaction vectors~\cite{huang2024copa} to ground manipulation behaviors. 
However, these methods remain limited for fine-grained control due to the sparsity of language constraints and the ambiguity inherent in applying 2D grounding to complex 3D scenes.
To overcome this, VLAs~\cite{brohan2022rt,zitkovich2023rt,zhen20243d,black2025pi_,zhao2025cot,li2025coa} directly co-fine-tune large language models with continuous robot trajectories to output low-level action tokens. 
Similarly, video-based approaches~\cite{du2023learning,bharadhwajgen2act,zhao2025taste,patel2025robotic,liang2025dreamitate,zhen2025tesseract,bharadhwaj2024track2act} employ video generation or prediction networks to synthesize future states from human or robot demonstrations, deriving action-level supervision from these visual dynamics. 
To ground these dynamics in 3D, most recent studies PointWorld~\cite{huang2026pointworld} and FlowDreamer~\cite{guo2026flowdreamer} directly predict point-cloud flow for robot and object motion. However, scaling such 3D world models remains constrained by the scarcity of high-quality 3D manipulation data compared to 2D video.
In parallel to these paradigms, SuSIE~\cite{blackzero} leverages image-editing diffusion models (e.g., InstructPix2Pix~\cite{brooks2023instructpix2pix}) to synthesize subgoal images, which then serve as visual guidance for a goal-conditioned policy. 
Unlike SuSIE's purely 2D formulation, this work explicitly grounds visual editing priors into inter-object 3D transformations, providing a more robust spatial foundation for open-world manipulation.
While the concurrent work GoalVLA~\cite{chen2025goal} also utilizes generative subgoals, it decouples the scaling factors of the active and passive objects during alignment. This inconsistent scale estimation fails to maintain global scene geometry, leading to significant spatial offsets. In contrast, our method enforces a unified scale constraint during 3D registration, ensuring the structural integrity of the edited scene for high-precision manipulation.

\begin{figure*}[htbp]
    \centering
    \includegraphics[width=0.8\linewidth]{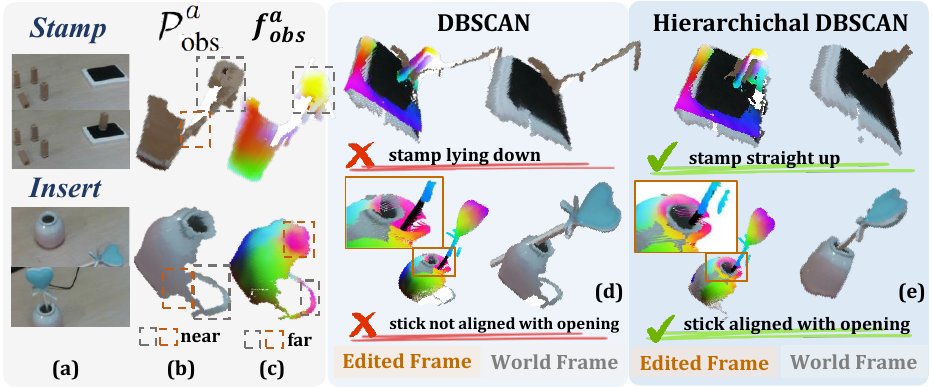}
    \caption{\textbf{Illustration of the 2D-3D hierarchical point-cloud filtering}. Colorful points in block (c) and (d-e) represent $\mathcal{P_{\text{obs}}}$ and $\mathcal{P_{\text{edit}}}$ with DINO features visualized via PCA, respectively. \textbf{(a) Task}: observed (top) and edited (bottom) images for \textit{stamping} and \textit{insertion}. \textbf{(b) Spatial space}: flying-edge points (gray boxes) of the stamp and vase are spatially proximal to valid points (orange boxes). \textbf{(c) Feature space}: flying-edge points (gray boxes) are distant from valid points (orange boxes) with similar PCA colors. \textbf{(d) Spatial clustering}: it fails when the stamp in horizontal or the stick is misaligned with the vase opening. \textbf{(e) Hierarchical filtering}: it successfully removes flying-edge points and recovers the correct spatial alignment.}
    \label{fig: method_filter}
\end{figure*}

\section{Method}
\label{sec:method}

At the core of our approach lies a simple question: can image editing provide stronger spatial priors for manipulation?
Edited images implicitly specify how objects should move and relate spatially.
This insight motivates us to formulate manipulation as predicting the inter-object 3D transformations (Sec.~\ref{sec:method_formulation}) and design a perception–reasoning–execution framework that converts visual edits into executable trajectories.

\noindent \textbf{Overview.}
An overview of our pipeline is illustrated in Fig.~\ref{fig:pipeline}. 
In the perception stage, we extract 3D spatial priors from the edited image to ground the high-level intent (Sec.~\ref{sec:method_prior}).  
In the reasoning stage, we propose a noise-robust cross-state point cloud registration for real-world settings, enabling reliable estimation of the 3D inter-object transformation via the edited state (Sec.~\ref{sec:method_registration}).
Finally in the execution stage, the estimated transformation is converted into the target pose to optimize the end-effector trajectory (Sec.~\ref{sec:method_execution}). 

\subsection{Task Formulation from an Editing Perspective}
\label{sec:method_formulation}
We formulate robotic manipulation as predicting the relative transformation of objects via visual editing.
Given an initial RGB-D observation $(I_{\text{obs}}, D)$ and a free-form language instruction $\mathcal{L}$, 
our goal is to generate a 6-DoF end-effector trajectory $\tau$ that executes the intended manipulation. 
The instruction $\mathcal{L}$ specifies a subtask-level manipulation rather than a high-level long-horizon command. Complex tasks can be decomposed into subtasks using a high-level planner~\cite{song2023llm,hu2024look}.
Specifically, $\mathcal{L}$ describes either a single target object $\mathcal{O}_{a}$ to be manipulated 
(e.g., “open the red drawer”), or an interaction between an active and a passive object $(\mathcal{O}_{a}, \mathcal{O}_{p})$ (e.g., “cover the teapot with the lid”).  
Leveraging the inherent spatial reasoning embedded in image editing, we formulate each manipulation as 
predicting a target relative transformation $\mathbf{T}_a \in \text{SE}(3)$ of the active object $\mathcal{O}_{a}$, mapping it from the observed state to the edited state. 
\subsection{Spatial Prior Extraction from Editing}
\label{sec:method_prior}

Given the current RGB observation $I_{\text{obs}} \in \mathbb{R}^{H \times W \times 3}$ and a task description $\mathcal{L}$, 
we generate an edited image $I_{\text{edit}} \in \mathbb{R}^{H \times W \times 3}$ conditioned on $\mathcal{L}$ using modern image-editing models~\cite{comanici2025gemini, wu2025qwen} to depict the target post-manipulation state of the active object $\mathcal{O}_a$ visually.
To recover its geometry,  
we lift $I_{\text{edit}}$ into a pixel-aligned point cloud $\mathcal{P}_{\text{edit}} \in \mathbb{R}^{(H \times W) \times 3}$ using a monocular depth estimator (e.g., VGGT~\cite{wang2025vggt}).
However, resolution mismatch between $I_{\text{edit}}$ and the depth estimator may cause spatial detail loss if directly processed.
To mitigate this, we extract binary masks $\mathcal{M}_{\text{edit}}^a$ and $\mathcal{M}_{\text{edit}}^p$ of $\mathcal{O}_a$ and $\mathcal{O}_p$ from $I_{\text{edit}}$, 
using LLMDet~\cite{fu2025llmdet} for language-grounded localization and SAM~\cite{kirillov2023segment} for pixel-level refinement.
For single-object instructions (e.g., ``open the red drawer"), the passive object $\mathcal{O}_p$ denotes its functionally coupled static surroundings (e.g., the drawer housing).
We then crop the tight bounding box enclosing $\mathcal{M}_{\text{edit}}^a$ and $\mathcal{M}_{\text{edit}}^p$, and resize or pad it by the original $I_{\text{edit}}$ to match the estimator's input resolution.
For resized images, the predicted depth is upsampled back to the cropped RGB resolution, ensuring one-to-one pixel correspondence.
This preserves spatial detail and yields accurate 3D grounding of manipulated regions in $\mathcal{M}_{\text{edit}}^a \cup \mathcal{M}_{\text{edit}}^p$ in $I_{\text{edit}}$.  

\subsection{Cross-state Point Cloud Registration}
\label{sec:method_registration}
To estimate the 6-DoF transformation $\mathbf{T}_a$ of the active object $\mathcal{O}_a$, we register current and edited point clouds. 
While registration is well-studied in reconstruction~\cite{zeng20173dmatch}, applying across edited states is challenging: observations are noisy and incomplete (Fig.~\ref{fig: method_filter}(b)),  and interacting objects $(\mathcal{O}_a, \mathcal{O}_p)$ may move, deform or occlude each other (Fig.~\ref{fig: method_filter}(a)). 
To handle these issues, we propose a cross-state registration pipeline that sequentially filters unreliable points, performs object-centric alignment, and applies unified scale correction to maintain consistent spatial reasoning.


\noindent\textbf{Point Cloud Filtering.}
RGB-D sensors often produce floating edge points due to depth discontinuities and sensor blur (Fig.~\ref{fig: method_filter}(b)).
Such artifacts degrade the accuracy of registration, especially for scale-sensitive manipulation.
Classic density-based filters (e.g., DBSCAN~\cite{schubert2017dbscan}) may fail to remove them, because these artifacts remain locally dense and close to valid regions (Fig.~\ref{fig: method_filter}(b)). 
Even depth-refinement methods~\cite{lin2025prompting} still output spatially coherent flying points once lifted into 3D.
We observe that, while these flying-edge points are spatially adjacent to valid points, they are far from inliers with similar visual features (Fig.~\ref{fig: method_filter}(c)).
To exploit this, we extract 2D features via DINOv3~\cite{simeoni2025dinov3} and cluster them via K-Means to group pixels with similar appearance.
DBSCAN is then applied within each cluster to remove spatial outliers (intra-cluster filtering), followed by refinement across clusters (inter-cluster filtering) (Fig.~\ref{fig: method_filter}(e)).
This hierarchical 2D-3D fused filtering suppresses boundary artifacts and stabilizes downstream registration.

\noindent\textbf{Point Cloud Registration.}
We separately register the observed point clouds of the active and passive objects ($\mathcal{P}_{\text{obs}}^a$ and $\mathcal{P}_{\text{obs}}^p$) to the frames of their edited counterparts ($\mathcal{P}_{\text{edit}}^a$ and $\mathcal{P}_{\text{edit}}^p$). The pixel-aligned point clouds are defined as:
\begin{equation} 
\begin{aligned}
\mathcal{P}_{\text{obs}}^{p/a} 
&= \{\,\mathbf{p}_i^{\text{obs}} \in \mathbb{R}^3 \mid i \in \mathcal{M}_{\text{obs}}^{p/a}\,\}, \\
\mathcal{P}_{\text{edit}}^{p/a} 
&= \{\,\mathbf{p}_i^{\text{edit}} \in \mathbb{R}^3 \mid i \in \mathcal{M}_{\text{edit}}^{p/a}\,\},
\end{aligned}
\end{equation}
where $i$ indexes pixels and the superscript $p/a$ denotes the passive or active object.
A fundamental challenge lies in establishing reliable correspondences $\mathcal{C}^{p/a}$.
Traditional registration~\cite{wang2019deep,rusu2009fast} or multi-view matching~\cite{lowe2004distinctive, sarlin2020superglue} methods assumes geometric and appearance consistency, which breaks between current and edited states.
The active object $\mathcal{O}_a$ may move, deform (e.g., ``open the red drawer"), interact with $\mathcal{O}_p$, or become occluded (e.g., ``insert the toast into the toaster"), leading to sparse and ambiguous matches.
In contrast, image editing inherently preserves the same viewpoint and pixel-level consistency for static regions (including $\mathcal{O}_p$). Therefore for $\mathcal{O}_p$ we form dense pixel-to-pixel correspondence:  
\begin{equation}
\begin{aligned}
\mathcal{C}^p &= \bigl\{ (\mathbf{p}_i^\text{obs}, \mathbf{p}_i^\text{edit}) \,\big|\, i \in \mathcal{M}_{\text{obs}}^p \cap \mathcal{M}_{\text{edit}}^p \bigr\},
\end{aligned}
\end{equation}
where each observed point $\mathbf{p}_i^{\text{obs}}$ is directly paired with its corresponding edited point $\mathbf{p}_i^{\text{edit}}$ in the intersection of the two masks.
For $\mathcal{O}_a$, we use semantic features $f$ from DINOv3~\cite{simeoni2025dinov3} to extract point correspondences. Unlike geometric or low-level features, semantic features encode object-level identity and remain robust to occlusion, partial observations, and deformation. 
For each edited point $\mathbf{p}_j^{\text{edit}}$, we find its nearest neighbor in $\mathcal{P}_{\text{obs}}^a$ by cosine feature distance $\mathrm{dist}(\cdot, \cdot)$, filtering out pairs whose distance exceeds a threshold $d_{\text{thr}}=0.3$:
\begin{equation}
\mathcal{C}^a = \left\{ (\mathbf{p}_{i^*}^{\text{obs}}, \mathbf{p}_j^{\text{edit}})\ \middle|\ 
\begin{aligned}
 & i^* = \arg\min_{i \in \mathcal{M}_{\text{obs}}^a} \, \mathrm{dist}(f_{i}^{\text{obs}}, f_j^{\mathrm{edit}}), \\
 & j \in \mathcal{M}_{\text{edit}}^a, \ 
   \mathrm{dist}(\mathbf{p}_{i^*}^{\text{obs}} - \mathbf{p}_j^{\text{edit}}) < d_{\text{thr}}
\end{aligned}
\right\}.
\end{equation}
With $\mathcal{C}^a$ and $\mathcal{C}^p$ we estimate the transformation for each object using the Umeyama algorithm~\cite{Umeyama}, solving for rotation $\mathbf{R}_{a/p}\in \text{SO}(3)$, translation $\mathbf{t}_{a/p}\in \mathbb{R}^3$ and scale  $s_{a/p}\in \mathbb{R}_{+}$:
\begin{equation}
s_{a/p}\cdot \mathbf{R}_{a/p} \mathcal{P}_{\text{obs}}^{a/p} + \mathbf{t}_{a/p} \approx \mathcal{P}_{\text{edit}}^{a/p}.
\end{equation}

\begin{figure}[h]
    \centering
    \includegraphics[width=0.8\linewidth]{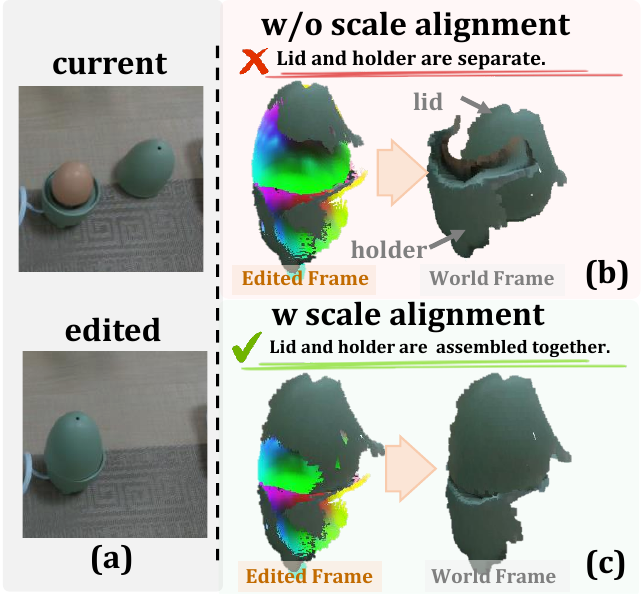}
    \caption{\textbf{Illustration of scale alignment}. Colorful points are $\mathcal{P_{\text{edit}}}$ with DINO features after PCA, while green points are $\mathcal{P_{\text{obs}}}$. \textbf{(a)}: Observation and edited image of task ``cover the lid onto the holder''.  \textbf{(b) Without alignment}: the two parts (lid and holder) drift apart in the world frame when transforming back under different scale from the edited frame. \textbf{(c) With alignment}: enforcing a consistent scale maintains the stable spatial relationship between the parts when transformed back to the world frame.}
    \label{fig: method_scale}
\end{figure}

\noindent\textbf{Relative Transformation Computation.}
Although the registration yields two reasonable transformations for $\mathcal{O}^a$ and $\mathcal{O}^p$, they are estimated under potentially different scales ($s_a\neq s_b$).
When transformed back to the world frame, this scale inconsistency causes noticeable offsets (Fig.~\ref{fig: method_scale}(b)), which can impair precise manipulation.
To ensure consistent scaling, we take the passive object as reference, since its pixel-to-pixel registration provides a relatively accurate scale mapping between the observed and edited coordinate frames.
We thus set $s_a=s_p$ and recompute the active object's rotation $\mathbf{R}_{a}$ and translation $\mathbf{t}_{a}$ to align both objects under a unified scale (Fig.~\ref{fig: method_scale}(c)).
Notably, the original scale gap is actually small (typically $<$0.5), further suggesting that image editing preserves strong spatial coherence across states.
To obtain the final world-frame transformation $\mathbf{T}_a$ of the active object, we first compute its scale-free relative transformation with respect to the passive object in the observation frame $[\mathbf{R}_{a|p}^{o}\mid\mathbf{t}_{a|p}^{o}]$:
\begin{equation}
 \begin{aligned}
 \mathbf{R}_{a|p}^{o} &= \mathbf{R}_p^{-1}\mathbf{R}_a, \\
 \mathbf{t}_{a|p}^{o} &= \mathbf{R}_p^{-1}(\mathbf{t}_a/s_a-\mathbf{t}_p/{s_p}).
 \end{aligned}
\end{equation}
We transform this relative motion into the world frame using the observation-to-world transformation $[\mathbf{R}_{\text{o2w}}\mid\mathbf{t}_{\text{o2w}}]$:
\begin{equation}
    \begin{aligned}
    \mathbf{R}_{a|p}^{\text{w}} &= \mathbf{R}_{\text{o2w}}\mathbf{R}_{a|p}^{o}\mathbf{R}_{\text{o2w}}^T, \\
    \mathbf{t}_{a|p}^{\text{w}} &= \mathbf{R}_{\text{o2w}}\mathbf{t}_{a|p}^{o}+\mathbf{t}_{\text{o2w}}-\mathbf{R}_{a|p}^{\text{w}}\mathbf{t}_{\text{o2w}}.
    \end{aligned}
\end{equation}
Finally, the active object's transformation in the world frame is given by $\mathbf{T}_a=[\mathbf{R}_{a|p}^{\text{w}}\mid\mathbf{t}_{a|p}^{\text{w}}]$.


\subsection{Edited Goal Informed Execution}
\label{sec:method_execution}
\begin{figure}[h]
    \centering
    \includegraphics[width=0.75\linewidth]{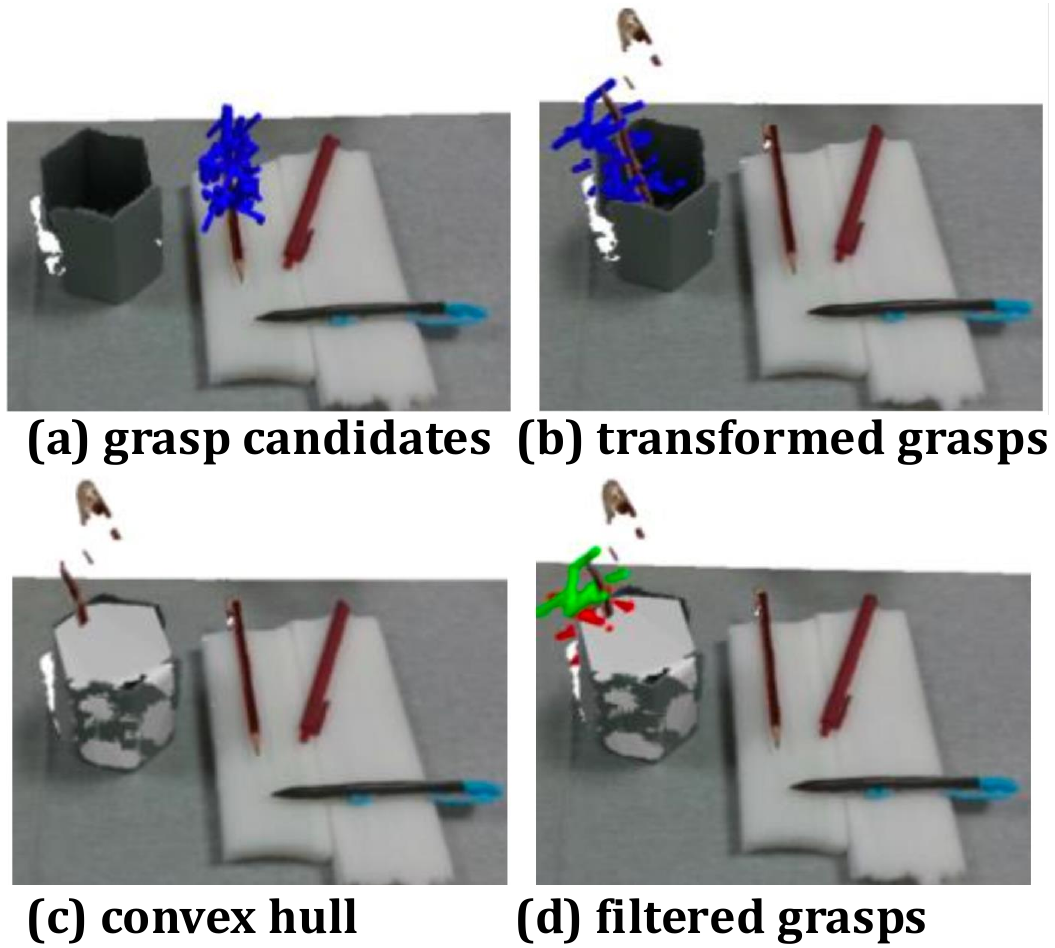}
    \caption{\textbf{Edited-informed grasping}. \textbf{(a) Candidate grasps} (blue) generated by AnyGrasp on the observed point cloud of the pencil. \textbf{(b) Transformed grasps} (blue) derived from the candidate set using the edit-informed transformation.  \textbf{(c) Collision convex hull} (gray mesh) of the holder. \textbf{(d) Filtered grasps}: red grasps indicate collisions with the holder, while green grasps denote valid task-specific candidates.}
    \label{fig: method_grasp}
    \vspace{-2em}
\end{figure}

To translate the predicted transformation into executable robot motions, we decouple the manipulation task into two sequential stages: grasping and transformation. 
While off-the-shelf grasping generators like AnyGrasp~\cite{fang2023anygrasp} can produce numerous grasp candidates for a target object, they are often task-agnostic. 
For example, ``insert the pen from the tip into the holder" requires grasping the pen from its top or body, not its tip, to avoid future collision with the holder.
The edited goal offers a strong task-specific spatial prior for feasibility. 
Specifically, we compute the convex hull of the passive object's point cloud (Fig.~\ref{fig: method_grasp}(c)). 
For each candidate grasp $\mathcal{G}$, we compute its corresponding pose at the goal state by applying the estimated transformation $\mathbf{T}_a$ (assuming the gripper and active object remain rigidly attached pose-grasp) (Fig.~\ref{fig: method_grasp}(a)(b)). Any grasp that results in a collision between the gripper and the passive object's convex hull in the edited state is discarded, thus retaining only task-feasible grasps (Fig.~\ref{fig: method_grasp}(d)). Finally, we employ CuRobo~\cite{sundaralingam2023curobo} for motion planning, utilizing environment voxels to ensure collision-free execution throughout the trajectory.

\section{Experiment}
\label{sec:eval}

In this section, we evaluate and analyze \codename to address three key questions:
\textbf{(1)} How well does our image-editing-based zero-shot registration perform in aligning manipulation pairs (Sec.~\ref{sec: exp_pcd_register})?
\textbf{(2)} To what extent can our editing-based manipulation framework generalize in open-world scenarios (Sec.~\ref{sec: exp_open_manip})?
\textbf{(3)} Can the image-editing prior support robust and long-horizon manipulation (Sec.~\ref{sec: exp_long_horizon})?

\subsection{Point Cloud Registration for Manipulation}
\label{sec: exp_pcd_register}
\noindent\textbf{Tasks.} 
To evaluate our registration method on manipulation pairs, we collect real-world scenes captured using a single-view RGB-D camera, as no existing 
one meets our needs. 
Each scene contains an active object $\mathcal{O}_a$, a passive object $\mathcal{O}_p$, and a natural language instruction describing the interaction. Given the two partial point clouds and the instruction, the task is to predict the relative 6-DoF transformation of $\mathcal{O}_a$ with respect to $\mathcal{O}_p$ that fulfill the described manipulation. 
Collected pairs covers diverse manipulation types (e.g., insertion, covering, placing, assembling, cutting). For quantitative evaluation, we scan object meshes via AR Code. Ground-truth transformations are derived by estimating the poses from pre-collected RGB-D human demonstrations using FoundationPose~\cite{wen2024foundationpose}.
Performance is measured using Root Mean Squared Error (RMSE) of rotation and translation.
\begin{figure}[h]
    \centering
    \includegraphics[width=\linewidth]{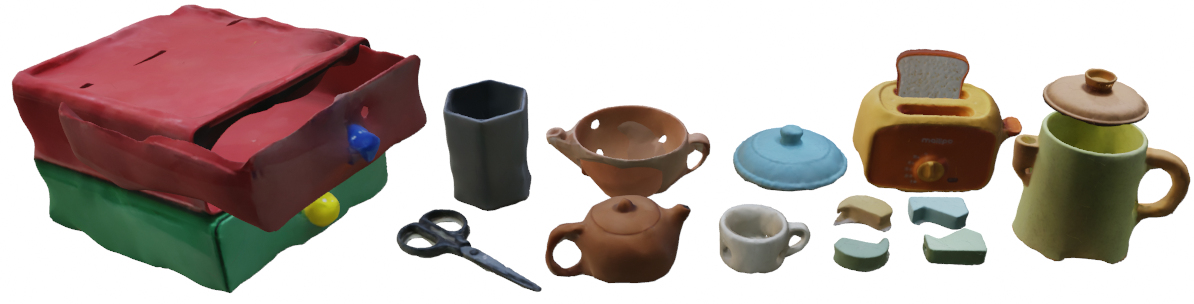}
    \caption{\textbf{Mesh of objects scanned by AR-Code App.} }
    \label{fig:exp_pcd_mesh}
\end{figure}

\noindent\textbf{Baselines.} We compare our method with two point cloud-based methods: 
1) \textbf{Two by Two (2BY2)}~\cite{qi2025two}, which predicts relative transformations between two object point clouds via a two-step $\text{SE}(3)$ pose-estimation pipeline for multi-task assembly, 
2) \textbf{AnyPlace}~\cite{zhao2025anyplace}, which predict placement poses from local point clouds cropped at VLM-proposed locations.

\begin{figure*}[h]
    \centering
    \setlength{\abovecaptionskip}{2pt}
    \includegraphics[width=\linewidth]{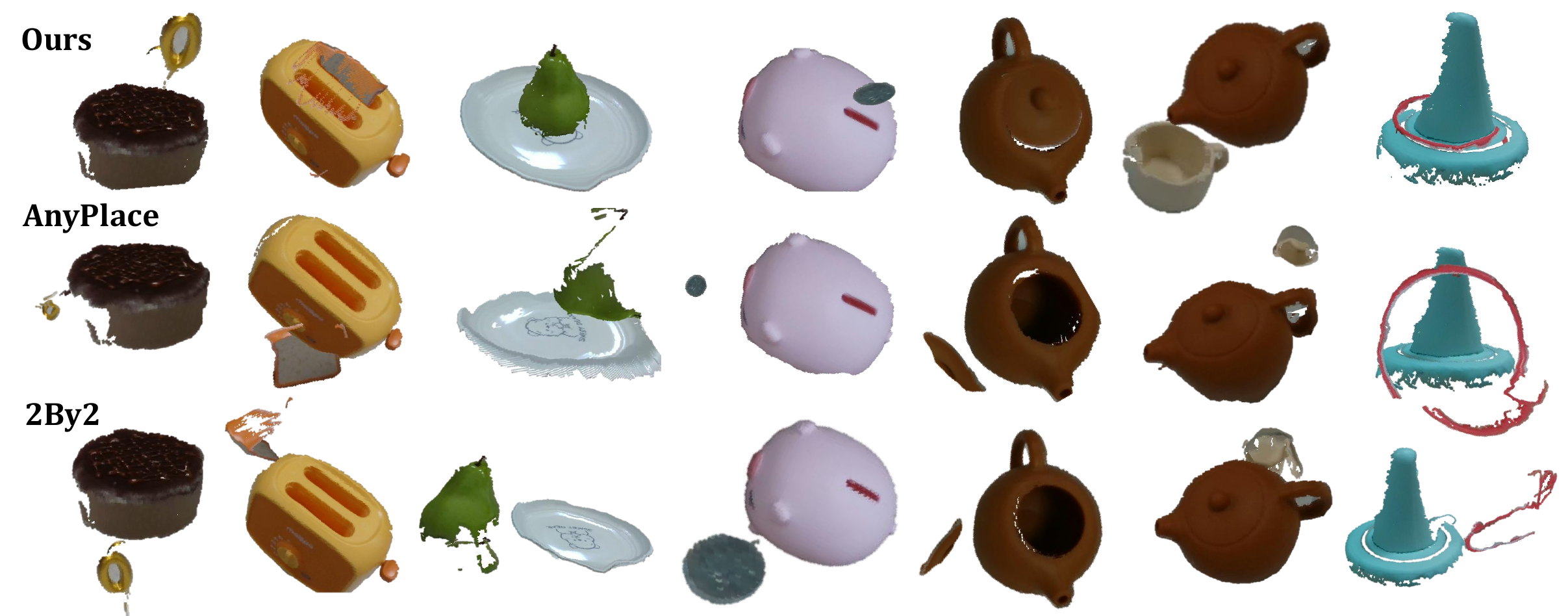}
    \caption{\textbf{Qualitative results of point cloud registration across diverse manipulation tasks.}
    \codename consistently aligns active and passive objects under various task configurations, showcasing strong generalization and robustness to noisy, partial real-world point clouds.}
    \label{fig:exp_pcd_regist}
\vspace{-1em}
\end{figure*}

\begin{table*}[h]
\centering
\caption{\textbf{Quantitative results of point cloud registration.}}
\vspace{-0.5em}
\resizebox{0.8\linewidth}{!}{\begin{tabular}{c;{1pt/1pt}cl;{1pt/1pt}ll;{1pt/1pt}cc;{1pt/1pt}cc;{1pt/1pt}cc} \hline
        \multirow{2}{*}{\textbf{Tasks}}                & \multicolumn{2}{c;{1pt/1pt}}{Lid covering} & \multicolumn{2}{l;{1pt/1pt}}{Toast insertion}                 & \multicolumn{2}{l;{1pt/1pt}}{Block assembly} & \multicolumn{2}{c;{1pt/1pt}}{Tea pouring} & \multicolumn{2}{c}{Drawer opening}  \\ 
        \cline{2-11}
                                                       & 2BY2 & \multicolumn{1}{c;{1pt/1pt}}{Ours}  & \multicolumn{1}{c}{2BY2} & \multicolumn{1}{c;{1pt/1pt}}{Ours} & \multicolumn{1}{c}{2BY2} & Ours              & 2BY2 & Ours                               & 2BY2 & \multicolumn{1}{c}{Ours}     \\ 
        \hline
        \multicolumn{1}{l;{1pt/1pt}}{\textbf{RMSE(t)~$\downarrow$}} & 0.091 & \textbf{0.003} & 0.095 &  \textbf{0.015} & / & 0.005 & / &  0.014    &  / &  0.017      \\
        \multicolumn{1}{l;{1pt/1pt}}{\textbf{RMSE(R)~$\downarrow$}} & 16.54 & \textbf{8.736} & 35.12   &  \textbf{11.10} & / & 30.05& / & 21.53  &  / & 2.614          \\
        \hline
\end{tabular}}
\label{tab: exp_pcd_regist}
\end{table*}

\noindent\textbf{Results.}
As shown in Fig.~\ref{fig:exp_pcd_regist} qualitatively, \codename generalizes well across diverse manipulation tasks and is markedly more robust to noisy, partial point clouds than all baselines.
Compared to 2BY2 quantitatively in Tab.~\ref{tab: exp_pcd_regist}, \codename achieves lower translation and rotation RMSE despite not relying on mesh.
This advantage mainly stems from the strong spatial priors embedded in image-editing models. 
Our proposed reasoning mechanism further lifts these implicit 2D constraints into coherent 3D relationships, enabling reliable alignment under real-world noise and occlusions.
In contrast, both AnyPlace~\cite{zhao2025anyplace} and 2BY2~\cite{qi2025two} struggle to generalize across tasks.
AnyPlace is fine-tuned on point clouds from simulation environments for tasks such as insertion, stacking, hanging, and placing, while 2BY2 is trained on mesh-sampled point clouds for insertion, covering, and placing. 
Their dependence on clean, task-specific training data limits their transferability to noisy and incomplete real-world observations, leading to a noticeable generalization gap.
These results highlight that image-editing priors offer a strong and transferable spatial understanding that enables robust point cloud registration across unseen tasks and real-world variations.

\begin{table}
\centering
\caption{Success rate of 13 real-world manipulation tasks. `/' indicates the method is not applicable for that task.}
\vspace{-0.5em}
\label{tab: exp_open_success}
\resizebox{0.95\linewidth}{!}{
\begin{tabular}{lcccc} 
\hline
\textbf{Tasks}   & \textbf{VoxPoser} & \textbf{CoPa} & \textbf{Rekep} & \textbf{Ours}    \\ 
\hline
Egg placing      & 2/10              & 2/10          & 4/10           & \textbf{6/10}    \\
Coin insertion   & 0/10              & 0/10          & 0/10           & \textbf{5/10}    \\
Pencil insertion & 0/10              & 4/10          & 3/10           & \textbf{7/10}    \\
Toast insertion  & 0/10              & 0/10          & 0/10           & \textbf{6/10}    \\
Lid covering     & 0/10              & 3/10          & 4/10           & \textbf{8/10}    \\
Pen-cap covering & 0/10              & 1/10          & 2/10           & \textbf{6/10}    \\
Tea pouring      & 0/10              & 1/10          & 3/10           & \textbf{6/10}    \\
Toast cutting    & 0/10              & 0/10          & 5/10           & \textbf{8/10}    \\
Block assembly   & 0/10              & 1/10          & 0/10           & \textbf{6/10}    \\
Ring stacking    & 2/10              & 1/10          & 3/10           & \textbf{8/10}    \\ 
\hline
Total            & 4.0\%             & 13.0\%        & 24.0\%         & \textbf{66.0\%}  \\ 
\hline
Drawer opening   & 2/10              & 4/10          & /              & \textbf{6/10}    \\
Drawer closing   & 4/10              & 4/10          & /              & \textbf{7/10}    \\
Toaster opening  & 2/10              & 1/10          & /              & \textbf{5/10}    \\ 
\hline
Total            & 26.7\%            & 30.0\%        & /              & \textbf{60.0\%}  \\
\hline
\end{tabular}}
\vspace{-2.0em}
\end{table}

\begin{figure*}[h]
    \centering
    \setlength{\abovecaptionskip}{2pt}
    \includegraphics[width=1.0\linewidth]{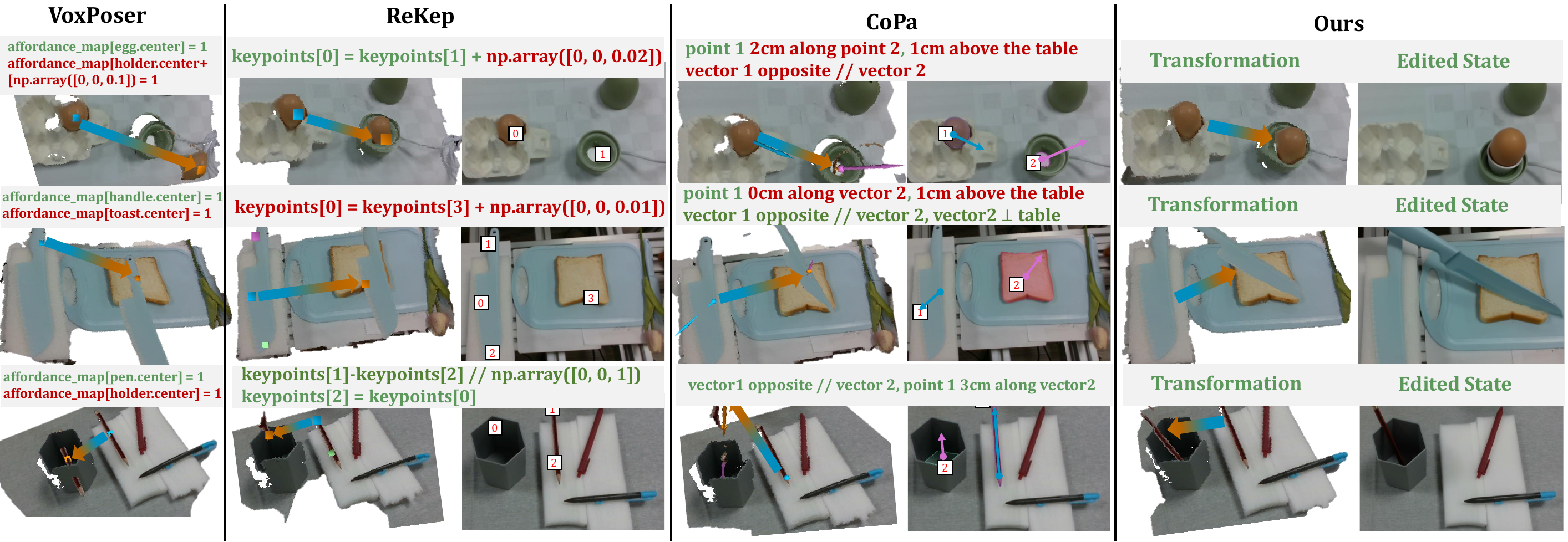}
    \caption{\textbf{Qualitative comparison of different manipulation representations.} The blue to orange arrows indicate the target manipulation pose. Voxposer~\cite{huang2023voxposer} grounds manipulation at the center of the object, ReKep~\cite{huang2025rekep} uses keypoints, CoPa~\cite{huang2024copa} uses keypoints and vectors, and our approaches uses a full 3D inter-object transformation. }
    \label{fig:exp_percept}
\vspace{-1em}
\end{figure*}

\begin{figure*}[h]
    \centering
    \setlength{\abovecaptionskip}{2pt}
    \includegraphics[width=1.0\linewidth]{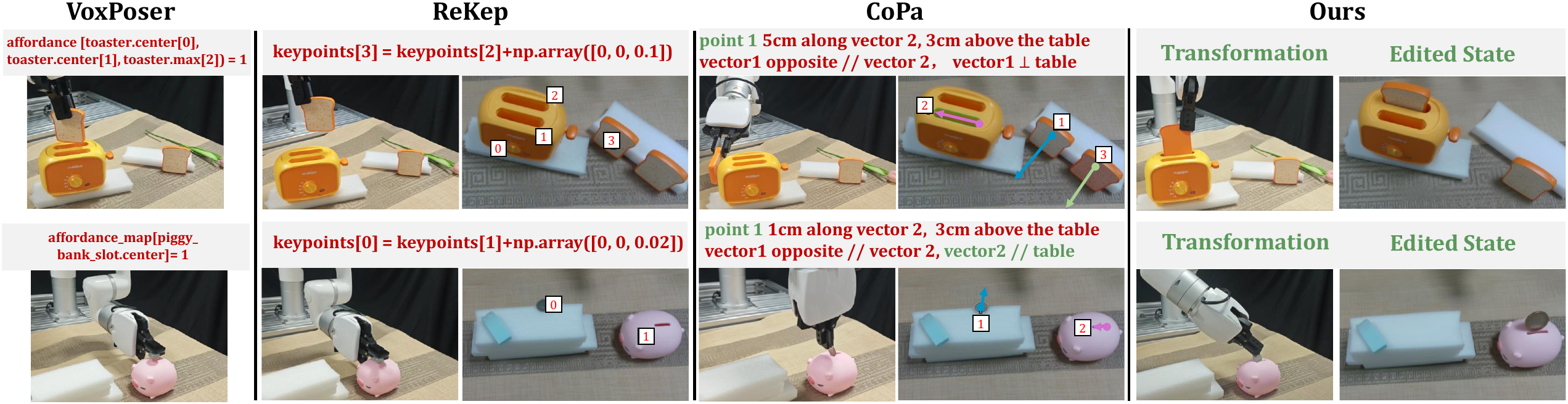}
    \caption{\textbf{Qualitative results on real-world insertion tasks} (\textit{Toast, Coin}). Voxposer~\cite{huang2023voxposer} fails to infer rotations; ReKep~\cite{huang2025rekep} misidentifies keypoints and rotations; CoPa~\cite{huang2024copa} cannot reliably capture vector constraints; our method recovers precise inter-object 3D transformations. }
    \label{fig:exp_real_world}
\vspace{-1em}
\end{figure*}

\begin{figure}[h]
    \centering
    \setlength{\abovecaptionskip}{2pt}
    \includegraphics[width=0.95\linewidth]{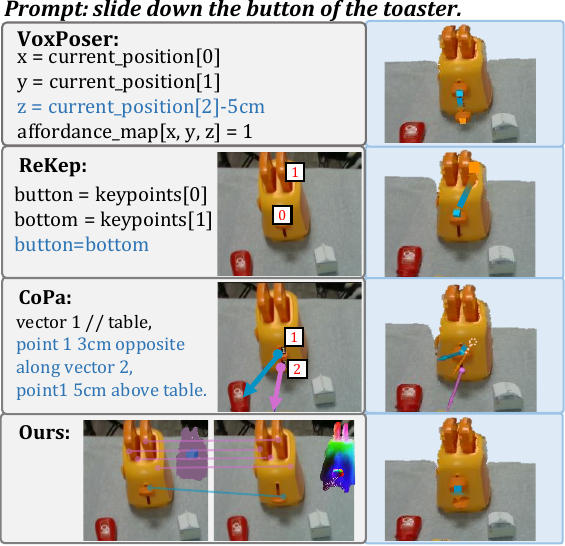}
    \caption{\textbf{Qualitative analysis of articulated manipulation.} }
    \label{fig:exp_articulate}
\vspace{-2.0em}
\end{figure}

\begin{figure*}
    \centering
    \setlength{\abovecaptionskip}{2pt}
    \includegraphics[width=0.95\linewidth]{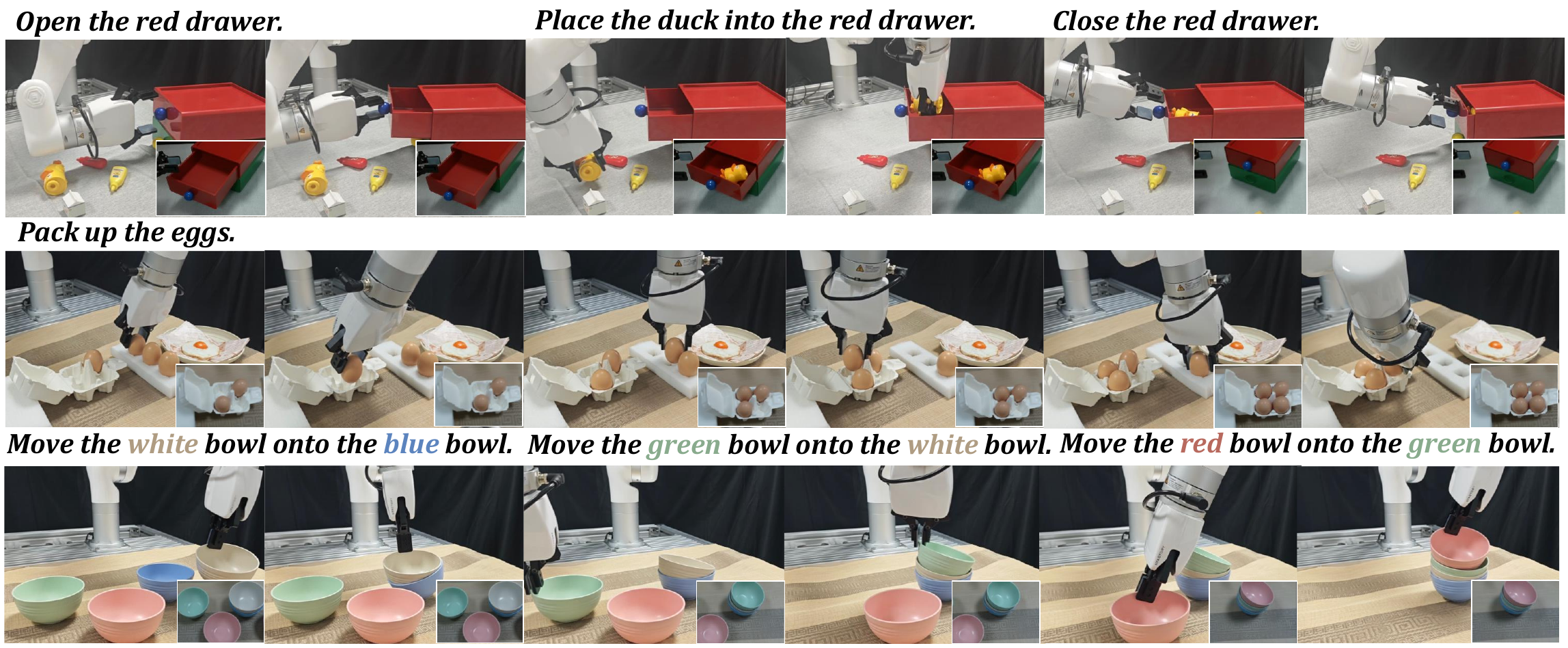}
    \caption{\textbf{Example rollouts of long-horizon manipulation tasks.} Bottom right corner shows the edited prior for each step.}
    \label{fig:exp_real_long}
\end{figure*}

\subsection{Open-world Manipulation}
\label{sec: exp_open_manip}
\noindent\textbf{Hardware Configuration.} 
Our experiments are conducted on a UFACTORY xArm7 robotic arm equipped with its UFACTORY xArm Gripper G2. An Intel RealSense D435i RGB-D camera is mounted opposite the robot to capture a third-person view of the workspace.

\noindent\textbf{Tasks and Metrics.}
We evaluate the open-world manipulation capability of \codename across a diverse set of everyday object-centric tasks, covering aspects from high-precision manipulation to articulated-object manipulation.
In total, we select 13 representative tasks, including egg placing, coin insertion, pencil insertion, toast insertion, lid covering, pen-cap covering, tea pouring, toast cutting, block assembly, ring stacking, drawer opening and closing, and toaster opening. 
Each task is executed for 10 trials with random object poses, and overall success rates are reported in Tab.~\ref{tab: exp_open_success} with more details in the appendix.

\noindent\textbf{Baselines.}
As analyzed in Sec.~\ref{sec: exp_pcd_register}, \textbf{Two by Two}~\cite{qi2025two} and \textbf{AnyPlace}~\cite{zhao2025anyplace} generalize poorly to single-camera, real-world manipulation setups. 
Therefore, we compare our method with three additional zero-shot open-world manipulation baselines: 
1) \textbf{Voxposer}~\cite{huang2023voxposer}, which uses LLM-generated code to build 3D value maps conditioned on language instructions for trajectory synthesis;
2) \textbf{CoPA}~\cite{huang2024copa}, which employs VLMs to infer spatial constraints between interaction keypoints and interaction surface vectors; and 
3) \textbf{Rekep}~\cite{huang2025rekep}, which formulates VLM-predicted relational keypoints as cost terms for trajectory optimization. 
We always provide CoPA with best available masks.

\noindent\textbf{Results.} 
\codename exhibits strong performance in task diversity, fine-grained manipulation, and execution robustness compared with baselines. 
This advantage stems from implicit 2D spatial cues in edited images, which are effectively lifted into 3D transformations through our object-centric formulation. 
Qualitative results in Fig.~\ref{fig:exp_percept} and Fig.~\ref{fig:exp_real_world} illustrate these strengths.
We analyze the performance from two perspectives: \textbf{the limits of language-based constraints} and \textbf{the challenges of input representations}.
Language-based constraints suffer from sparse and ambiguous 3D guidance, missing fine-grained relations (i.e., rotations, contact geometry, and object-to-object alignment) and thus leading to failures in precision tasks such as toast or coin insertion (Fig.~\ref{fig:exp_real_world}).
For geometry-sensitive tasks like egg placing or knife cutting (1st and 2nd row of Fig.~\ref{fig:exp_percept}), 
VoxPoser~\cite{huang2023voxposer} and ReKep~\cite{huang2025rekep} exhibit limited rotational awareness, while CoPa may produce contradictory constraints due to weak geometric understanding.
In contrast, our method uses edited images to provide implicit spatial priors that encode both the rotation and interaction regions of objects, enabling accurate 3D alignment even for fine-grained manipulation.
Beyond language limitations, the input modality itself also constraints performance.
VoxPoser can convert phrases like ``slide down" into z-axis motion (1st row of Fig.~\ref{fig:exp_articulate}) but cannot infer metric geometry without visual grounding (e.g., -5cm); 
ReKep may misidentify keypoints without task-specific keypoint extraction (e.g., misidentified ``bottom" keypoint in 2nd row of Fig.~\ref{fig:exp_articulate}).
CoPa projects 3D vectors onto 2D observation, making it  sensitive to noisy point clouds (e.g., incorrect surface normal of the button in 3rd row of Fig.~\ref{fig:exp_articulate}) and ambiguous shapes (i.e., ellipsoids like eggs in 1st row of Fig.~\ref{fig:exp_percept}). 
Our method leverages subject consistency and visual correspondence between current and the edited states (4th row in Fig.~\ref{fig:exp_articulate}), providing a robust global context that generalizes across diverse object geometries and articulated-object tasks.


\begin{figure}[h] 
    \centering
    \setlength{\abovecaptionskip}{2pt}
    \includegraphics[width=0.98\linewidth]{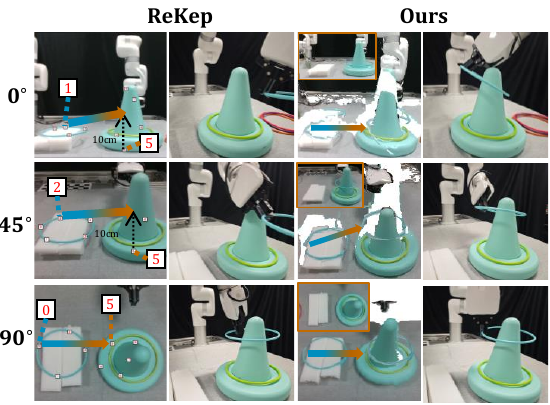}
    \caption{Qualitative analysis of camera‑viewpoint effects.}
    \label{fig:exp_abla_viewpoint}
\end{figure}

\begin{table}[h]
\centering
\vspace{-0.5em}
\setlength\tabcolsep{3pt} 
\caption{Quantative results of viewpoint influence (\textit{ring stacking}).}
\vspace{-0.5em}
\label{tab: exp_abla_view}
\resizebox{0.9\linewidth}{!}{
\begin{tabular}{cccccc} 
\hline
\begin{tabular}[c]{@{}c@{}}\textbf{View-}\\\textbf{point}\end{tabular} & \textbf{ReKep} & \textbf{CoPa} & \begin{tabular}[c]{@{}c@{}}\textbf{Ours}\\\textbf{(wo filter)}\end{tabular} & \begin{tabular}[c]{@{}c@{}}\textbf{Ours}\\\textbf{(wo scale)}\end{tabular} & \begin{tabular}[c]{@{}c@{}} \textbf{Ours}\\\textbf{(full)}\end{tabular}  \\ 
\hline
$0^\circ$                                                             &       1/10        & 1/10           &      2/10           &   3/10                      & \textbf{6/10}                                                  \\
$45^\circ$                                                            &      3/10         & 4/10           &         6/10        &   3/10                        & \textbf{8/10}                                                  \\
$90^\circ$                                                            &        2/10       &    6/10            &         7/10                                                         &  4/10                         & \textbf{8/10}                                                  \\
\hline
\end{tabular}}
\vspace{-1.0em}
\end{table}

\subsection{Long-horizon Manipulation}
\label{sec: exp_long_horizon}
Following the setup in Sec.~\ref{sec: exp_open_manip}, we further evaluate \codename on long-horizon manipulation tasks to demonstrate its understanding of multi-step object-centric interactions. 
Long-horizon tasks typically require decomposition into subtasks, where the execution of each step depends on the final state of the previous one.
We design three long-horizon tasks: putting a duck into the red drawer, packing the eggs, and setting up the table. 
Fig.~\ref{fig:exp_real_long} shows example execution rollouts, highlighting that \codename maintains accurate object alignment and successfully completes each subtask in sequence.
To further analyze the benefits of our approach, we compare the use of spatial priors from edited images against video generation priors. Edited-image priors exhibit stronger adherence to semantic constraints and better background consistency, resulting in more reliable and coherent long-horizon manipulation.



\subsection{Ablation Study}
We ablate our pipeline on the ring stacking task under three viewpoints ($0^\circ$, $45^\circ$, and $90^\circ$). 
Success rates over 10 trials are in Tab.~\ref{tab: exp_abla_view} and qualitative results are shown in Fig.~\ref{fig:exp_abla_viewpoint}.
Without our proposed point cloud filtering, the success rate at the $0^\circ$ viewpoint drops a lot, as the ring becomes almost line-like in the image, making depth highly unreliable (1st row in Fig.~\ref{fig:exp_abla_viewpoint}. 
Removing scale alignment also degrades performance, since stacking requires precise relative placement and inconsistent scales break this alignment.
Compared with baselines, our approach is notably more robust to viewpoint changes, as illustrated in Fig.~\ref{fig:exp_abla_viewpoint}. ReKep~\cite{huang2025rekep} and CoPa~\cite{huang2024copa} both rely on relationships between 2D keypoints or projected vectors, which are inherently limited by the field of view and depth accuracy of corresponding keypoints. 
In contrast, our method lifts the implicit spatial priors from edited images into full 3D transformations and performs dense registration, leading to greater resilience to noise, occlusion, and partial geometry. 

\vspace{-0.5em}
\section{Conclusion}
\label{sec:conclu}
In this work, we present \codename, a generalizable representation that lifts image editing as 3D priors to extract inter-object transformations. 
Leveraging implicit spatial cues in edited images, \codename provides precise 3D relational understanding, enabling robust generalization across viewpoints, object geometries, and fine-grained manipulation tasks.
This work marks a promising step toward scalable open-world manipulation.
Despite these promises, limitations remain. \codename currently handles rigid-body interactions and does not address soft-body or deformable-object manipulation.
The framework relies on motion planning to execute and thus tasks requiring intermediate trajectories may need additional motion priors or task-specific planning heuristics. As with most language-based models, it requires moderate prompt engineering to ensure consistent edits.

\clearpage
\setcounter{page}{1}
\maketitlesupplementary

\section{Implementation Details}
\subsection{Pseudo-code for Hierarchical Point Cloud Filtering}
As shown in Algo.~\ref{algo: pcd_filter}, given the object point cloud $\mathcal{P}^{a/p}_{\text{obs}}$ projected from the current RGB-D observation, the corresponding DINO feature $\mathbf{F}_{\text{obs}}$ and the object mask $\mathcal{M}^{a/p}_{\text{obs}}$, the algorithm outputs a filtered set of valid 3D points along with a pixel-aligned binary mask indicating the retained regions.

\subsection{Implementation Details for Point Cloud Registration}
Since the DINO feature-based matching for the active object requires KNN to compute the distance matrix, we use the \texttt{cuml} library to accelerate the computation.


\subsection{Prompt for Image-Editing}
We use Qwen-Image-Edit and Gemini 2.5 Flash Image (Nano Banana) as our editing models. The prompts used for each task in open-world manipulation are provided in Tab.~\ref{tab: exp_edit_prompt}.

\begin{table}[h]
\centering
\caption{\textbf{A list of 13 open-world manipulation tasks}. We provide the prompt used to generate the edited image in our experiment.}
\label{tab: exp_edit_prompt}
\resizebox{\linewidth}{!}{
\begin{tabular}{ll} 
\hline
\textit{Egg placing}      & move the egg onto the green holder                                                                                  \\
\textit{Coin insertion}   & insert the coin into the piggy bank                                                                                 \\
\textit{Pencil insertion} & insert the pencil into the holder                                                                                   \\
\textit{Toast insertion}  & insert the toast into the toaster                                                                                   \\
\textit{Lid covering}     & move the lid onto the teapot                                                                                        \\
\textit{Pen-cap covering} & cover the pen with the pen cap                                                                                      \\
\textit{Tea pouring}      & teapot pours into the cup                                                                                           \\
\textit{Toast cutting}    & cut the toast with the knife                                                                                        \\
\textit{Block assembly}   & \begin{tabular}[c]{@{}l@{}}move the green block near the blue block \\so that their jagged edges meet\end{tabular}  \\
\textit{Ring stacking}    & toss the red ring over the base                                                                                     \\ 
\hline
\textit{Drawer opening}   & pull out the red drawer                                                                                             \\
\textit{Drawer closing}   & push the red drawer in                                                                                              \\
\textit{Toaster opening}  & move the slider of the toaster downwards                                                                            \\
\hline
\end{tabular}}
\end{table}

\SetKwComment{tcp}{\textcolor{blue!50!black}{$\triangleright$\ }}{\textcolor{blue!50!black}{}}

\begin{algorithm}[h]
	\caption{Hierarchical Point Cloud Filtering} 
	\label{algo: pcd_filter}
        \SetAlgoLined
        \DontPrintSemicolon
	\SetKwInput{KwInput}{Input}                
	\SetKwInput{KwOutput}{Output}              
	\KwInput{object point cloud $\mathcal{P}^{a/p}_{\text{obs}}\in \mathbb{R}^{M\times 3}$,
    DINO features of the point cloud $\mathbf{F}^{a/p}_{\text{obs}}\in \mathbb{R}^{M\times D}$,    Number of K-Means layers $K$,
    DBSCAN params $(\varepsilon, \text{MinPts})$}
	\KwOut{filtered point cloud $\mathcal{P'}^{a/p}_{\text{obs}}\in \mathbb{R}^{N \times 3}$, corresponding mask $\mathcal{M'}^{a/p}_{\text{obs}}\in \mathbb{B}^{M}$ of chosen area}
\tcp*[l]{\textcolor{blue!50!black}{Initialization}}
    $\mathbf{L} \gets -1 \in\mathbb{Z}^{M \times 1}$;
    $\text{gid}\gets 0$

\tcp*[l]{\textcolor{blue!50!black}{Stage 1: Feature Scaling}}
$\mathbf{F} \gets \text{Standardize}(\mathbf{\tilde F})$;

\tcp*[l]{\textcolor{blue!50!black}{Stage 2: Intra-cluster Filtering}}
$\mathbf{L_{\mathbf{\tilde F}}} \gets \text{KMeans}(\tilde{\mathbf{F}}, K)$; \tcp*[l]{Feature Layering}
\For{$k = 0$ \KwTo $K-1$}
{
    $\mathcal{I}_k \gets \{i \mid \mathbf{L_{\mathbf{\tilde F}}}[i] = k\}$
    
    \If{$|\mathcal{I}_k|<\text{MinPts}$}{\textbf{continue}}

    $\mathcal{P}^{a/p}_k=\mathcal{P}^{a/p}_{\text{obs}}[\mathcal{I}_k]$

    $\mathbf{Y}_k \gets \text{DBSCAN}(\mathcal{P}_k; \varepsilon, \text{MinPts})$

   Let $s_c$ be the size of cluster $c\neq -1$ \;
    \If{no valid cluster}{
        \textbf{continue}
    }

    $c^\star \gets \arg\max_c s_c$ \tcp*{dominant DBSCAN cluster}

    \If{$s_{c^\star} \ge S_{\min}$}{
        Assign global cluster: \\
        \Indp 
            $\mathcal{J} = \{ i \in \mathcal{I}_k \mid \mathbf{Y}_{k}[i] = c^\star \}$ \\
            $\mathbf{L}[i] \gets$ \text{gid} \quad $\forall i \in \mathcal{J}$ \\
        \Indm
        $\text{gid} \gets \text{gid}+ 1$
    }
}

$\mathcal{M}^{a/p}_{\text{intra}} \gets \mathbf{L} \neq -1$

\tcp*[l]{\textcolor{blue!50!black}{Stage 3: Inter-cluster Filtering}}
$\mathcal{P}^{a/p}_{\text{intra}} \gets \mathcal{P}^{a/p}_{\text{obs}}[\mathcal{M}^{a/p}_{\text{intra}}]$

$\mathbf{Y}_{\text{intra}} \gets \text{DBSCAN}(\mathcal{P}^{a/p}_{\text{intra}})$;

Let $s_c$ be cluster sizes for all $c\neq -1$
    
$c^\star \gets \arg\max_c s_c$ 

$\mathcal{M}^{a/p}_{\text{inter}}\gets \mathbf{Y}_{\text{inter}}=c^\star$\;
$\mathcal{M'}^{a/p}_{\text{obs}}\gets \mathcal{M}^{a/p}_{\text{intra}};\mathcal{M'}^{a/p}_{\text{obs}}[\mathcal{M}^{a/p}_{\text{inter}}] =\mathcal{M}^{a/p}_{\text{inter}} $

\Return $\mathcal{P'}^{a/p}_{\text{obs}}\gets \mathcal{P}^{a/p}_{\text{intra}}[\mathcal{M}^{a/p}_{\text{inter}}]$ and $\mathcal{M'}^{a/p}_{\text{obs}}$
\end{algorithm}

\section{More Visualization Results}
\begin{figure*}
    \centering
    \setlength{\abovecaptionskip}{2pt}
    \includegraphics[width=1.0\linewidth]{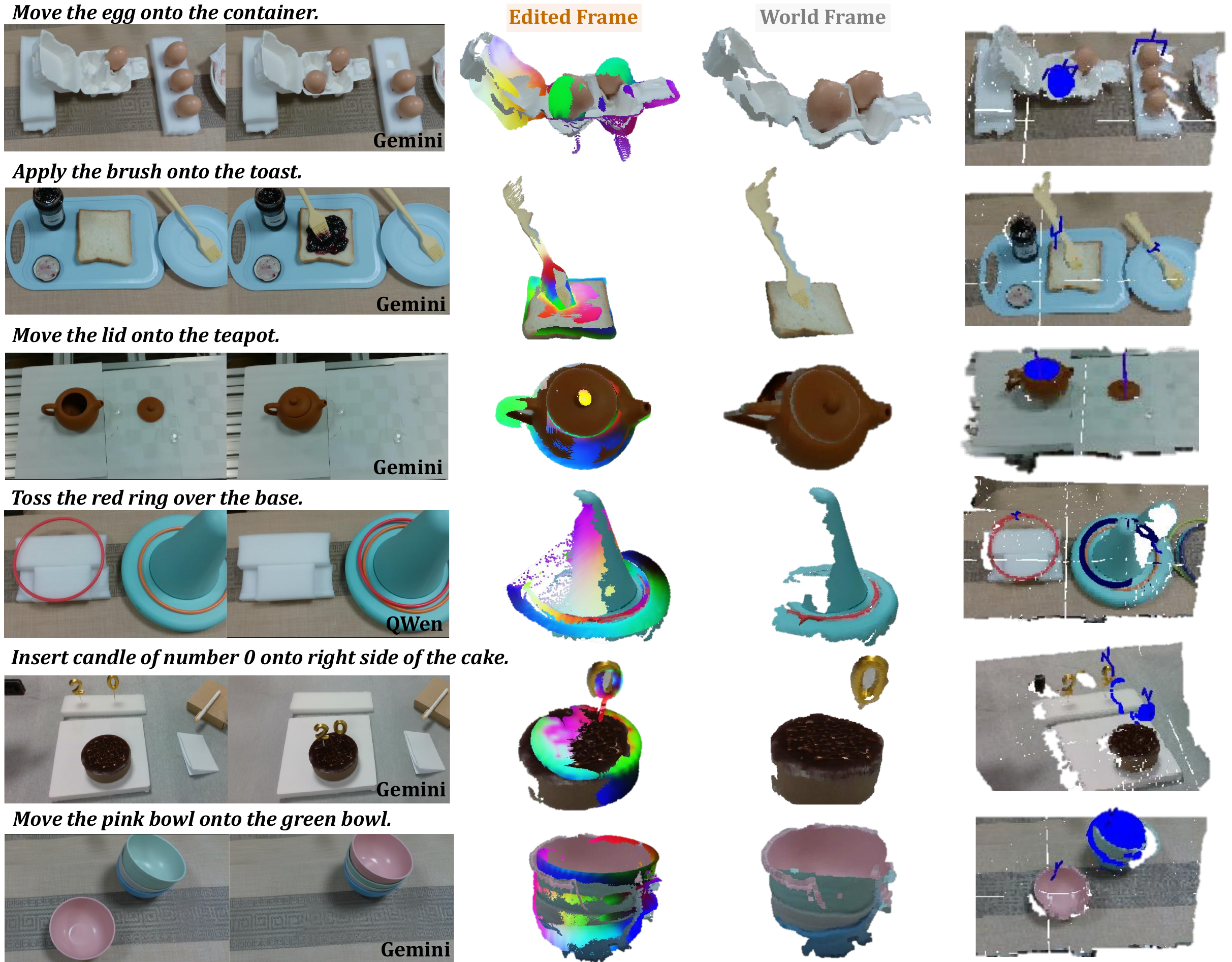}
    \caption{\textbf{More visualization results.} The first and second columns show the original observation and the edited state. The third and fourth columns show the registered point clouds in the edited frame and the world frame. The colored point clouds are $\mathcal{P}^{a/p}_{\text{edit}}$ after PCA. The last column shows the filtered grasp and the transformed active object.}
    \label{fig:supple_edit_prior}
\vspace{-1em}
\end{figure*}

More visualization results for cross-state point cloud registration and edited-informed grasping are shown in Fig.~\ref{fig:supple_edit_prior}, 

\begin{figure*}
    \centering
    \setlength{\abovecaptionskip}{2pt}
    \includegraphics[width=1.0\linewidth]{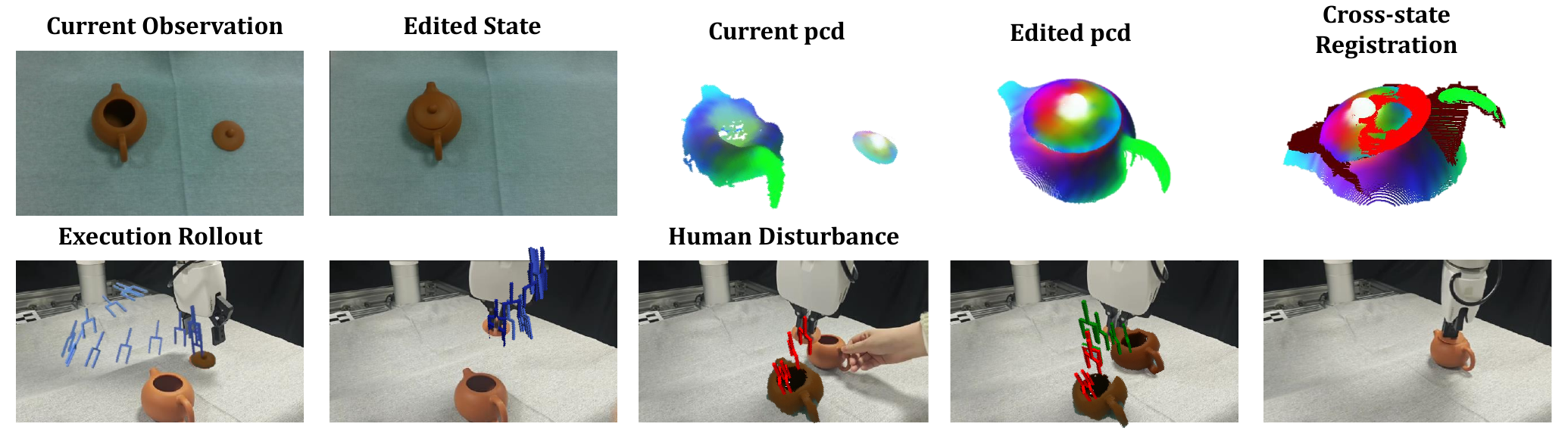}
    \caption{\textbf{Visualization of closed-loop execution rollout.}}
    \label{fig:supple_closed_loop}
\vspace{-1em}
\end{figure*}

\section{Closed-loop Manipulation}
To further demonstrate how our extracted 3D priors support closed-loop manipulation, we evaluate \codename on the \textit{Lid covering} task under human-induced disturbances, where the passive object is moved during execution (Fig.~\ref{fig:supple_closed_loop}).
We use Cutie~\cite{cheng2024cutie} to track the mask of the active object $\mathcal{O}_a$ across frames.
A straightforward approach is to track keypoints~\cite{xiao2024spatialtracker, xiao2025spatialtrackerv2} inside the mask to obtain point-to-point correspondences, but current keypoint trackers are insufficiently accurate, particularly under rotation, resulting in unreliable 3D alignment.
In contrast, dense pixel-wise matching with DINO features provides robust correspondences, enabling a more precise estimation of the active object's transformation for closed-loop control.

\section{Runtime Profiling}

\begin{figure}[h]
    \centering
    \includegraphics[width=\linewidth]{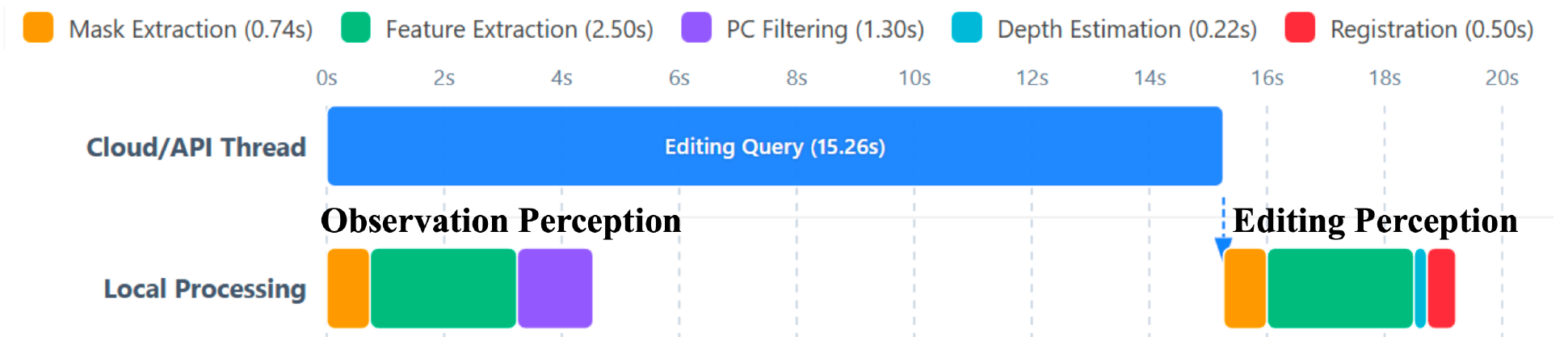}
    \caption{\textbf{Runtime Profiling.}}
    \label{fig:profiling}
\end{figure}

To analyze the computational overhead of our multi-module pipeline, we conducted runtime profiling as illustrated in Fig.~\ref{fig:profiling}. Adhering to a 'think-before-act' paradigm, computationally intensive modules are executed outside the primary control loop. Consequently, while perception remains efficient, the overall latency is primarily dominated by the image editing querying phase.

\section{System Error Breakdown}

\begin{figure}[h]
    \centering
    \includegraphics[width=\linewidth]{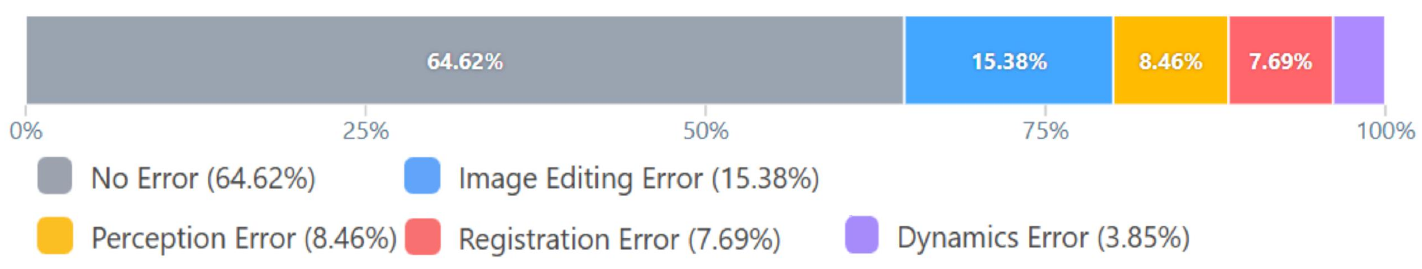}
    \caption{\textbf{System error breakdown.}}
    \label{fig:exp_sys_error}
\end{figure}

We conduct an empirical investigation of system errors by analyzing the failure cases from Tab.~\ref{tab: exp_open_success}. As illustrated in Fig.~\ref{fig:exp_sys_error}, the majority of failures are attributed to the image editing module. These cases typically involve unintended modifications to task-irrelevant scene elements or a failure to reflect the requested edits in the output. 
Perception and registration errors constitute another significant portion. These failures are predominantly triggered by small-scale objects or severe viewpoint occlusions, both of which hinder accurate spatial reasoning. In contrast, the low-level controller contributes only a minimal fraction, indicating that once a valid plan is generated, the execution remains relatively robust.

\section{More Results for Ablation}
\subsection{Comparisons between Image Editing Model and Video Generative Model}
Video generation is another potential approach to provideing 3D priors for manipulation.
In our comparison, we use Kling 1.6 and Veo 3 to generate video sequences conditioned on the same current observation, as shown in Fig.~\ref{fig:supple_edit_abla}.
However, compared with video generation, our priors from edited-images exhibit stronger adherence to semantic constraints and better subject consistency, resulting in more reliable and coherent long-horizon manipulation.

\subsection{Ablation on Different Editing Models}
\begin{figure*}[h]
    \centering
    \setlength{\abovecaptionskip}{2pt}
    \includegraphics[width=1.0\linewidth]{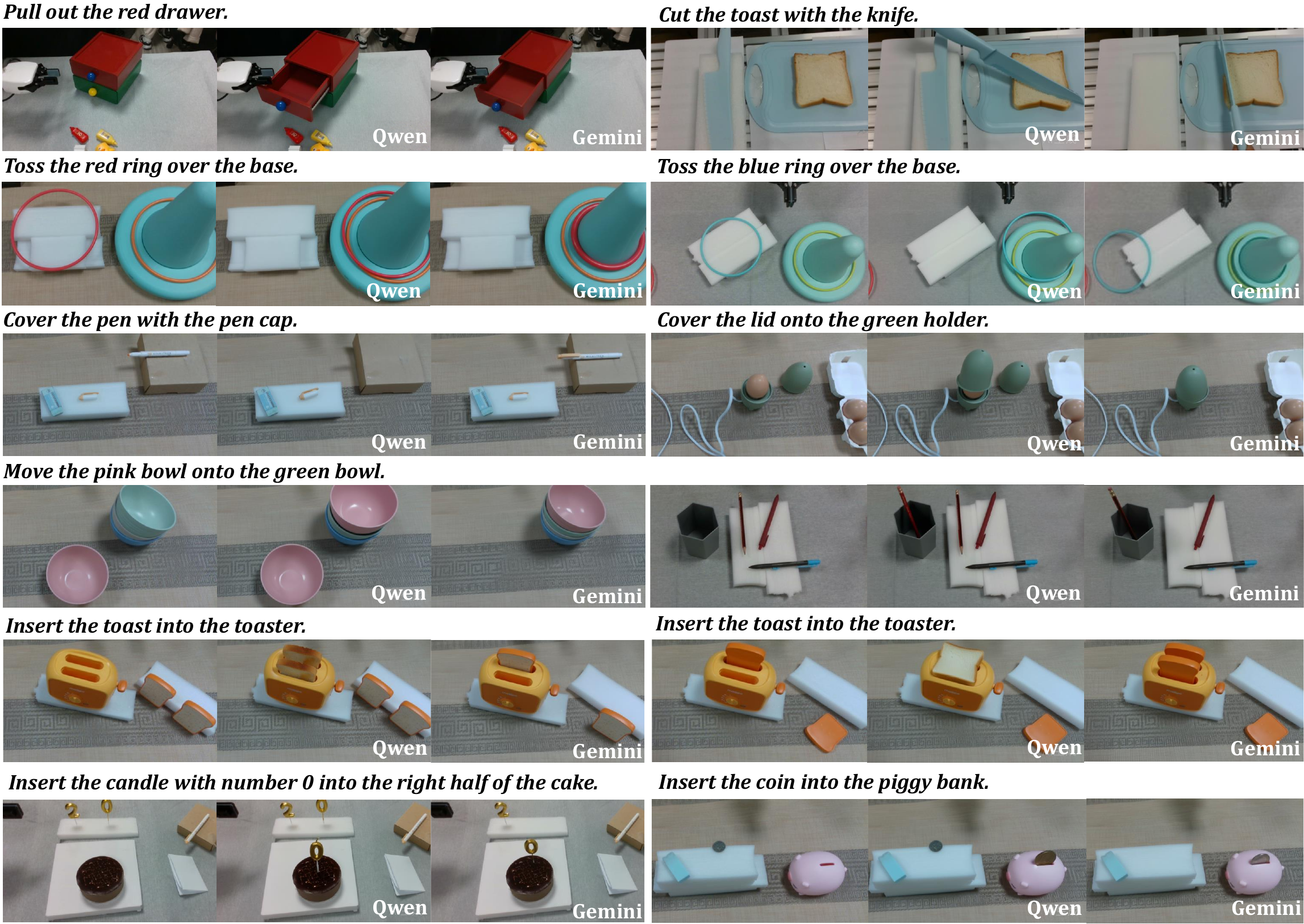}
    \caption{\textbf{Comparison of edited manipulation state with different editing models.} }
    \label{fig:supple_edit_abla}
\vspace{-1em}
\end{figure*}

We further ablate \codename in open world manipulation by comparing editing models QWen-Image-Edit and Gemini 2.5 Flash in Tab.~\ref{tab: exp_abla_edit}.
The edited priors are shown in Fig.~\ref{fig:supple_edit_abla}.
Gemini 2.5 Flash demonstrates stronger subject consistency and better adherence to semantic constraints.
However, it performs poorly in certain tool-use scenarios such as \textit{Ring stacking} and \textit{Toast cutting}.
QWen-Image-Edit, on the other hand, struggles with understanding directional relationships (e.g., in \textit{Candle insertion}) and shows limited scene awareness (e.g., \textit{Toast insertion}).
Besides, we observe that image editing does not always remove the active object from its original location.
To ensure correct extraction of the target priors for the active object, we perform a simple validation step by checking the overlap ratio between the extracted mask and the original object region.

\begin{table}
\centering
\caption{Ablation on open-world manipulation between different image editing models.}
\label{tab: exp_abla_edit}
\begin{tabular}{lcc} 
\hline
\textbf{Tasks}   & \textbf{Ours(QWen)} & \textbf{Ours(Gemini)}  \\ 
\hline
Egg placing      & \textbf{6/10}       & 5/10                  \\
Toast insertion  & 1/10                & \textbf{6/10}          \\
Pen-cap covering & 1/10                & \textbf{6/10}          \\
Toast cutting    & \textbf{8/10}       & 5/10                   \\
Ring stacking    & \textbf{8/10}       & 3/10                   \\
\hline
\end{tabular}
\end{table}

\section{More Results for Comparison}

While the concurrent work GoalVLA~\cite{chen2025goal} adopts a pipeline similar to ours, it overlooks the critical challenges of depth alignment and point cloud registration essential for fine-grained manipulation. 
This oversight leads to significantly lower success rates in precision-demanding tasks such as assembly and insertion, as quantified in Tab.~\ref{tab: exp_goalvla}. 
\begin{table}[h]
\centering
\caption{\textbf{Real-world comparison with GoalVLA~\cite{chen2025goal}.} Their neglect of scale consistency between active and passive objects throughout the pipeline results in significant spatial offsets. Consequently, their approach suffers from a remarkably low success rate in precision-demanding tasks such as fine-grained manipulation.}
\label{tab: exp_goalvla}
\resizebox{\linewidth}{!}{\begin{tabular}{lccccc} 
\hline
\textbf{Tasks} & Lid covering  & Pencil Insertion & Pen-cap covering & \multicolumn{1}{l}{Ring stacking} & \multicolumn{1}{l}{Drawer closing}  \\ 
\hline
GoalVLA        & 3/10          & 1/10             & 0/10             & 4/10                              & 1/10                                \\
\textbf{Ours(LAMP)}  & \textbf{8/10} & \textbf{7/10}    & \textbf{6/10}    & \textbf{8/10}                     & \textbf{7/10}                       \\
\hline
\end{tabular}}
\end{table}

\begin{figure}[htbp]
    \centering
    \includegraphics[width=\linewidth]{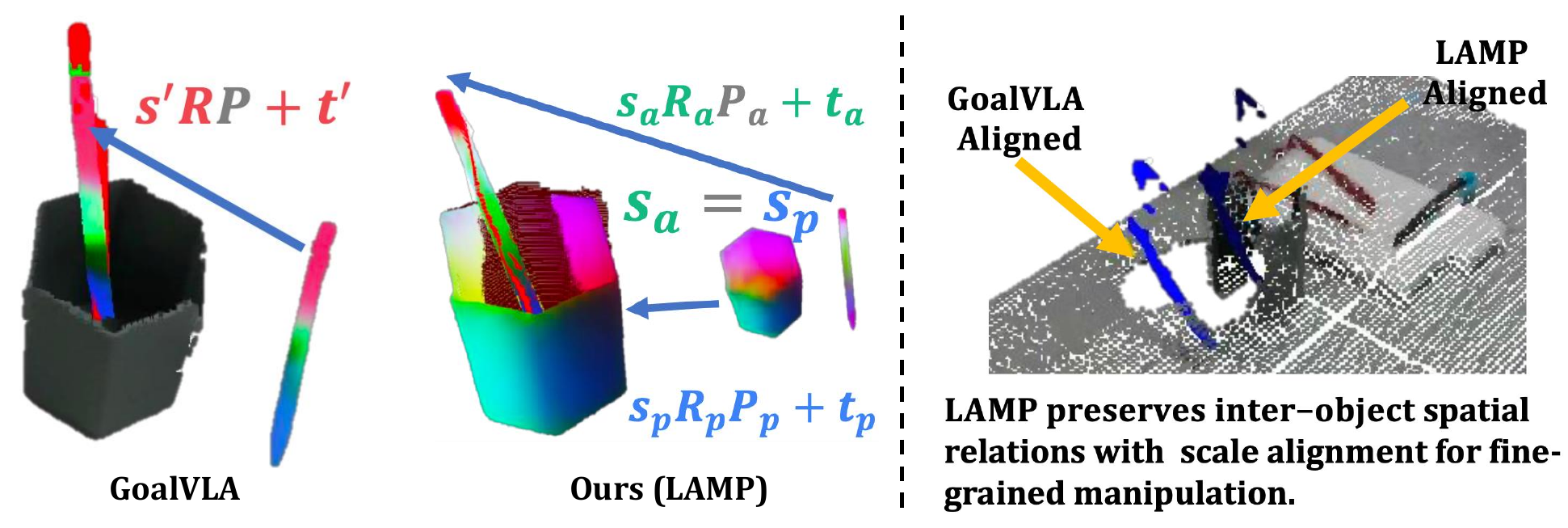}
    \caption{\textbf{Comparison of Alignment with GoalVLA.} While GoalVLA~\cite{chen2025goal} treats the passive object (e.g., the gray holder) as a static reference and aligns the active object (pencil) using an independent scale $s'$, this decoupling fails to account for global scene consistency. As illustrated in the right figure, such independent scaling distorts the relative spatial relationship between the two objects when transformed back to world coordinates. In contrast, our method enforces a unified scale across both the pencil and holder, strictly preserving their spatial configuration and ensuring the pencil remains correctly centered within the holder in 3D space.}
    \label{fig: goalvla_align_compare}
\end{figure}

\noindent We emphasize that achieving precise scale alignment between the edited and observed images is the key to lifting 2D edits into a reliable 3D prior for manipulation.
In our registration process, we enforce the constraint $s_{a} = s_{p}$ to ensure that when the objects are transformed back to world coordinates, the spatial relationship between the active and passive objects is strictly preserved.
\begin{equation}
s_{a/p}\cdot \mathbf{R}_{a/p} \mathcal{P}_{\text{obs}}^{a/p} + \mathbf{t}_{a/p} \approx \mathcal{P}_{\text{edit}}^{a/p}.
\end{equation}
In contrast, GoalVLA~\cite{chen2025goal} aligns edited images with observations via depth linear regression, computing the transformation of the active object under a optimized scale $s$.
Since physical objects are non-deformable, ignoring the consistency of $s$ and relying solely on $R$ and $T$ causes a shift in the relative spatial configuration.
While such offsets may be negligible for coarse tasks like pick-and-place as in their evaluations, even a 1\% scale error can result in significant translation offsets that are catastrophic for fine-grained manipulation as shown in Fig.~\ref{fig: goalvla_align_compare}.

\begin{figure}[htbp]
    \centering
    \includegraphics[width=\linewidth]{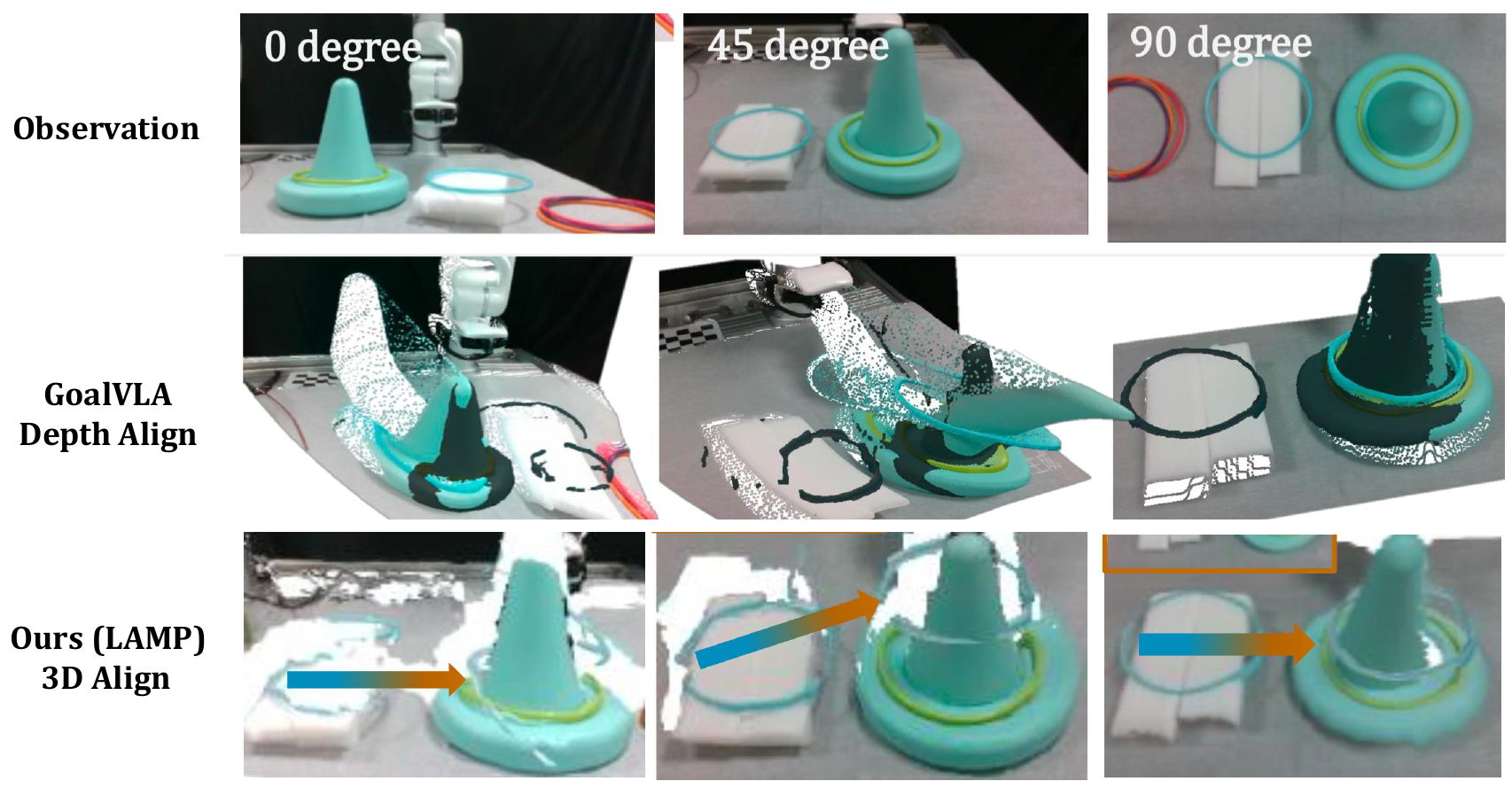}
    \caption{\textbf{Robustness comparison of viewpoint variation.} GoalVLA (row 2) exhibits noticeable scene shifts relative to the observation point cloud (dark) under different perspectives, our method (row 3) achieves stable 3D alignment. This demonstrates that our 3D-based registration is invariant to camera viewpoint changes, whereas 2D-based scale estimation is highly sensitive to perspective distortion.}
    \label{fig:goalvla_depth_compare}
\end{figure}

\noindent Besides, we evaluate the performance under varying camera viewpoints ($0^\circ$, $45^\circ$, and $90^\circ$) under the same edited image. As shown in Fig.~\ref{fig:goalvla_depth_compare}, GoalVLA's reliance on 2D depth linear regression (2nd row) leads to a significant scene shift relative to the observation. This is evident where the estimated point cloud of the edited image (colorized) drifts away from the observed point cloud (dark region) at $0^\circ$ and $45^\circ$. In contrast, our method (3rd row) performs registration directly in 3D space between the edited and world frames and demonstrates superior robustness to viewpoint changes.

\begin{figure*}
    \centering
    \setlength{\abovecaptionskip}{2pt}
    \includegraphics[width=1.0\linewidth]{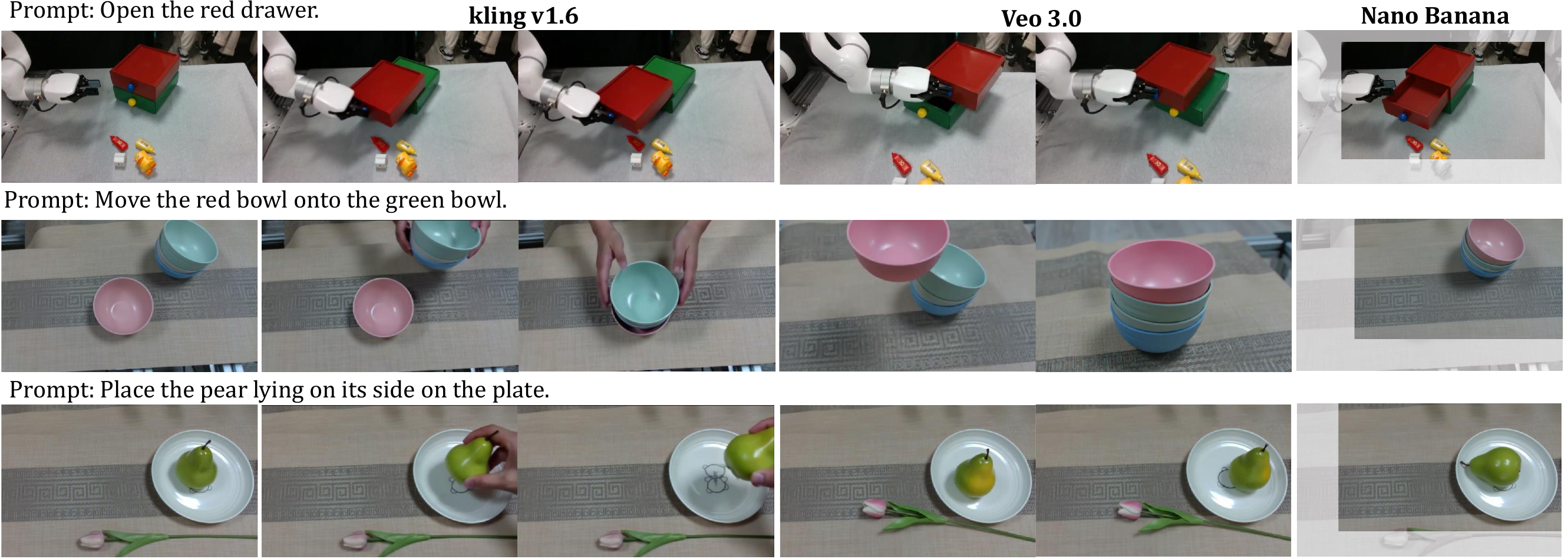}
    \caption{\textbf{Comparison between edited-image priors and video-generation priors for long-horizon manipulation.} 
    Edited-image priors provide stronger semantic adherence with better subject and background consistency.}
    \label{fig:exp_video_compare}
\vspace{-1em}
\end{figure*}

\section{Evaluation Details}
In this section, we provide the evaluation details for the evaluation section.
\subsection{Task Details for Point Cloud Registration}
To evaluate baselines that similar to our setting,  taking two point clouds and predicting the inter-object transformation, we collect real-world RGB-D observations covering a range of manipulation tasks, as described below.

\noindent\textbf{Candle insertion}: $\mathcal{O}_a$ is the candle and $\mathcal{O}_p$ is the cake. $\mathcal{L}$ refers to "insert the candle onto the cake". The goal is to insert the candle anywhere on the cake surface.

\noindent\textbf{Toast insertion}: $\mathcal{O}_a$ is the toast and $\mathcal{O}_p$ is the toaster.
$\mathcal{L}$ refers to "insert the toast into the toaster". The goal is to insert the toast into any valid slot of the toaster.

\noindent\textbf{Coin insertion}: $\mathcal{O}_a$ is the coin and $\mathcal{O}_p$ is the piggy bank. $\mathcal{L}$ refers to "insert the coin into the piggy bank". The goal is to align the coin with the bank's slot and orient it correctly for insertion.

\noindent\textbf{Pear placing}: $\mathcal{O}_a$ is the pear and $\mathcal{O}_p$ is the plate. $\mathcal{L}$ refers to "place the pear on the plate"
. The goal is to place the pear anywhere on the plate in any stable orientation.

\noindent\textbf{Lid covering}: $\mathcal{O}_a$ is the lid and $\mathcal{O}_p$ is the teapot. $\mathcal{L}$ refers to "cover the teapot with the lid". The goal is to place the lid onto the teapot opening with proper alignment.

\noindent\textbf{Tea pouring}: $\mathcal{O}_a$ is the teapot and $\mathcal{O}_p$ is the cup. $\mathcal{L}$ refers to "pour tea from the teapot into the cup". The goal is to rotate and position the teapot such that the spout aligns with and tilts over the cup.

\noindent\textbf{Ring stacking}: $\mathcal{O}_a$ is the ring and $\mathcal{O}_p$ is the base. $\mathcal{L}$ refers to "stack the ring onto the base". The goal is to align the ring hole with the peak of the base and then move the ring down to put them in place.

\noindent\textbf{Block assembly}: $\mathcal{O}_a$ is a block and $\mathcal{O}_p$ is another block or base structure. $\mathcal{L}$ refers to "assemble the two blocks together". The goal is to align their contact surfaces.




To ensure a fair comparison with the baselines, we train \textbf{Two by Two} using their official configurations and recenter all input point clouds at the origin for inference. For \textbf{AnyPlace} we directly evaluate using their publicly released pre-trained checkpoint.

\subsection{Details for Open-world Manipulation}
For each task, we rearrange the objects across 10  trials and ensure that they remain within the robot's reachable and kinematically feasible workspace.
To maintain identical initial configurations across baselines, we manually reset the scene after each execution.
Success rates are evaluated according to the task-specific criteria described below.

\noindent\textbf{Egg placing}: The environment includes an egg ($\mathcal{O}_a$) and an egg holder ($\mathcal{O}_p$), with task description $L$ "move the egg onto the green egg holder". The task involves grasping the egg, aligning it with the  holder and placing it stably onto the holder. The success criterion requires the egg resting upright on the holder without rolling or flipping.

\noindent\textbf{Coin insertion}: The environment includes a coin ($\mathcal{O}_a$) and a piggy bank ($\mathcal{O}_p$), with task description $L$ "insert the coin into the piggy bank". The task includes grasping the coin, aligning it with the slot of the piggy bank and inserting it into the slot. The success criterion requires successfully inserting the coin into the piggy bank through the slot.

\noindent\textbf{Pencil insertion}: The environment includes a pencil ($\mathcal{O}_a$) and a pencil holder ($\mathcal{O}_p$), with task description $L$ "insert the pencil into the holder". The task includes grasping the pencil, aligning it with the opening of the holder, and inserting it vertically into the holder.
The success criterion requires the pencil standing stably inside the holder.

\noindent\textbf{Toast insertion}: The environment includes a toast ($\mathcal{O}_a$) and a toaster ($\mathcal{O}_p$), with task description $L$ "insert the toast into the toaster". The task includes grasping the toast, aligning it with a toaster slot, and inserting it.
The success criterion requires the toast fully slid into a slot of the the toaster.

\noindent\textbf{Lid covering}: The environment includes a lid ($\mathcal{O}_a$) and a teapot ($\mathcal{O}_p$), with task description $L$ "cover the teapot with the lid". The task includes grasping the lid, aligning it with the teapot opening, and placing it.
The success criterion requires the lid fitting the teapot perfectly.

\noindent\textbf{Pen-cap covering}: The environment includes a pen cap ($\mathcal{O}_a$) and a pen body ($\mathcal{O}_p$), with task description $L$ "cover the pen with the pen cap". The task includes grasping the pen cap, aligning it with the pen tip of the pen body, and cover the pen cap onto the pen tip.
The success criterion requires the the pen cap fully attaching to the pen.

\noindent\textbf{Tea pouring}: The environment includes a teapot ($\mathcal{O}_a$) and a cup ($\mathcal{O}_p$), with task description $L$ "pour the tea from the teapot into the cup". The task includes grasping the teapot, tilt it over the cup, and maintaining the control.
The success criterion requires the the water visibly flowing into the teacup from the teapot.

\noindent\textbf{Toast cutting}: The environment includes a knife ($\mathcal{O}_a$) and a toast ($\mathcal{O}_p$), with task description $L$ "cut the toast with the knife". The task includes grasping the knife, aligning it with the toast, and cutting along a straight trajectory.
The success criterion requires a visible cut edge made through the toast.

\noindent\textbf{Block assembly}: The environment includes a block placed at the right hand ($\mathcal{O}_a$) and another matched block placed at the left hand($\mathcal{O}_p$), with task description $L$ "assemble the right block to the left block". The task includes grasping the right block, aligning it with the left block, and assembling it.
The success criterion requires a the block fitted correctly with another block.

\noindent\textbf{Ring stacking}: The environment includes a ring ($\mathcal{O}_a$) and base with peak ($\mathcal{O}_p$), with task description $L$ "insert the ring onto the base". The task includes grasping the ring, aligning it with the peak of the base, and lowering it.
The success criterion requires the ring fully placed onto the peak of the base.

\noindent\textbf{Drawer opening}: The environment includes a red drawer with handle ($\mathcal{O}_a$) (the drawer frame as $\mathcal{O}_p$), with task description $L$ "open the red drawer". The task includes grasping the handle and pulling the drawer outward along its rail direction.
The success criterion requires the drawer opens beyond a predefined threshold (i.e., 10cm).

\noindent\textbf{Drawer closing}: The environment includes the same drawer as \textbf{Drawer opening} but initially open, with task description $L$ "close the red drawer". The task includes pushing the opened drawer along its rail.
The success criterion requires the drawer opens pushed within a predefined threshold (i.e., 2cm).

\noindent\textbf{Toaster opening}: The environment includes the slide button of a toaster ($\mathcal{O}_a$) (the rest part of the toaster as $\mathcal{O}_p$), with task description $L$ "slide the button of the toaster downwards". The task includes sliding down the button of the toaster along its rail.
The success criterion requires the button of the toaster slid steadily and completely down.

To ensure a fail comparison with the baselines, we focus primarily on the interaction between the objects. 
Given the same RGB-D observations, we use GPT-4o to extract manipulation constraints for \textbf{VoxPoser}, \textbf{ReKep}, and \textbf{CoPa}. 
For all baselines, we provide the best available object masks and identical task instructions.
Since accurate keypoint localization in the real world depends heavily on the point cloud quality, our qualitative real-world comparisons in the main paper assume that each baseline is given the correct keypoint locations to better isolate the robustness of the extracted constraints.

\subsection{Details for Long-horizon Manipulation}
For the long-horizon manipulation, we design three tasks as detailed below.

\noindent\textbf{Putting a duck into the red drawer}: The environment contains stacked drawers (the red drawer on top of the green one), along with a duck and other toys on the table. The task consists of three stages: (i) opening the red drawer, (ii) placing the duck inside, and (iii) closing the red drawer.

\noindent\textbf{Packing up the eggs}: The environment contains three eggs standing upright in a row and an egg container with one egg already packed. The task consists of three stages, each involving picking up an egg from the row and placing it into an available slot in the container.

\noindent\textbf{Setting up the table}: The environment contains four bowls on the table colored with white, blue, red and green. The task requires stacking them sequentially according to the color order specified in the instruction.
{
    \small
    \bibliographystyle{ieeenat_fullname}
    \bibliography{main}
}


\end{document}